\documentclass[11pt]{article}
\usepackage{preprint}

\usepackage{graphicx}
\usepackage{booktabs}
\usepackage{multirow}
\usepackage{array}
\usepackage{xspace}
\usepackage{enumitem}
\usepackage[htt]{hyphenat}
\usepackage[numbers,sort&compress]{natbib}
\usepackage{xcolor}
\usepackage[colorlinks,linkcolor=blue!55!black,citecolor=blue!55!black,urlcolor=blue!55!black]{hyperref}

\newcommand{\gandr}{\textsc{GANDR}\xspace}

\title{\textsc{GANDR}: Claim Auditing for Verifiable Legal Answer Generation}

\author{%
  {\large\bfseries Chen Qian$^{1}$ \quad Yimeng Wang$^{1}$ \quad Yu Chen$^{2}$ \quad
  Lingfei Wu$^{2}$ \quad Andreas Stathopoulos$^{1}$}\\[0.5em]
  $^{1}$William \& Mary \quad $^{2}$Anytime AI\\[0.3em]
  \texttt{\{cqian03,ywang139,axstat\}@wm.edu} \quad
  \texttt{\{ychen,lwu\}@anytime-ai.com}%
}
\date{}


\begin{document}
\maketitle

\begin{abstract}
In high-stakes domains such as legal practice, a language-model answer is only useful to the extent that a reader can verify each claim against the source the system cites. Current grounded-generation pipelines score the answer as a whole, so a correct conclusion can rest on fabricated or loosely matched citations and still score well. Closing this gap requires both a system built for per-claim verification and an evaluation that measures it. We introduce \gandr (Grounded ANswer DRafter), a two-agent system in which a Drafter writes an answer in a structured legal-reasoning format and a separate Critic, with the same view as a human verifier, audits each claim against its cited source and emits a per-claim audit trace on every round. We pair it with a strict correctness criterion requiring every citation to resolve to a passage the retriever returned. On a $185$-item legal benchmark where all six systems share one backbone, one retrieval surface, and one citation instruction, \gandr ranks first on every primary metric, reaching $70.8\,\%$ strict accuracy and leading the strongest baseline by $11.3$ points ($p<0.01$). Reverting the protocol-anchored commit rule lowers strict accuracy by $22.7$ points, and the strict lead stays positive on three further backbones, at $+3.2$ to $+6.5$ points. This lead traces to the Drafter configuration and the protocol-anchored commit, not to rewriting. Against two law-trained annotators the audit flags under-supported claims at F1 $0.84$ as a binary detector, while its four-way verdict labels agree only weakly and are advisory. Code is available upon request.
\end{abstract}

\section{Introduction}
\label{sec:intro}

When a lawyer uses an AI assistant to research a legal question, the work that follows the answer is verification, not exploration~\citep{qian2026thinkingjustifyingaligninghighstakes}. The reader checks each claim against the source the system cites, asking whether that source states the rule and supports the argument. A single fabricated citation in a legal memo carries real professional cost. Audits put numbers on the failure mode: general-purpose LLMs hallucinate on $58\,\%$ to $88\,\%$ of open-domain legal queries~\citep{dahl2024legalfictions}, and even retrieval-grounded commercial legal-AI products still do so on $17\,\%$ to $43\,\%$~\citep{magesh2024hallucinationfree}.

Evaluation today, however, scores the answer in aggregate: pipelines place source passages in context, instruct the model to cite them, and report a single faithfulness rate, citation-precision score, or correctness verdict~\citep{liu2023ragfaithfulness,li2024citationhallucination,gao2023alce}, so a correct conclusion can rest on fabricated or loosely matched citations and still score well. A stricter metric alone is not enough: verifiability has to be built into the system, not patched on with a post-hoc score. Standard remedies do not close the gap. Single-LLM generation drifts on citations even when the disposition, the answer's operative conclusion, is right~\citep{niu2024ragtruth,liu2023ragfaithfulness,gao2023alce}. Stacking more LLM calls~\citep{shinn2023reflexion,madaan2023selfrefine,gou2024critic} need not reduce the slip rate, and in the default multi-agent configurations we test it rises rather than falls.

\begin{figure}[t]
\centering
\includegraphics[width=0.90\linewidth]{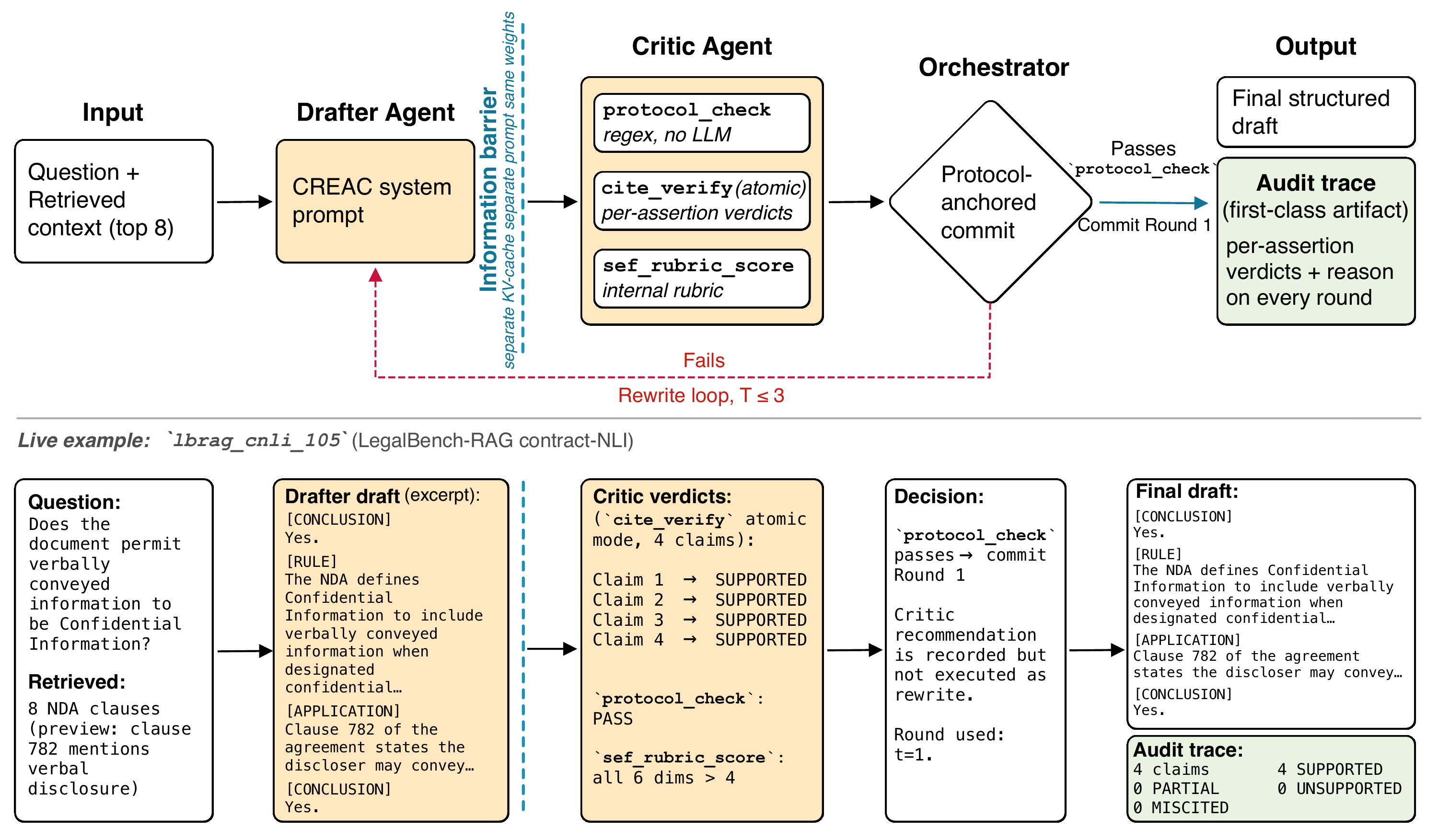}
\caption{\gandr architecture (top) and a live trace (bottom). Drafter and Critic share backbone weights but run in \emph{disjoint} LLM contexts: the Critic sees the question, retrieved passages, and the Drafter's final draft only, never the Drafter's chain of thought. The Critic invokes three tools (\texttt{protocol\_check}, atomic \texttt{cite\_verify}, \texttt{sef\_rubric\_score}) and writes per-assertion verdicts to the audit trace on every round. The orchestrator's commit rule is anchored to \texttt{protocol\_check}.}
\label{fig:architecture}
\end{figure}

We propose \gandr (Grounded ANswer DRafter), a two-agent system that builds per-claim verification into generation (Figure~\ref{fig:architecture}). A Drafter produces an answer in a structured legal-reasoning format (Conclusion, Rule, Explanation, Application, Conclusion)~\citep{qian2026thinkingjustifyingaligninghighstakes} that lays every assertion out in the order a verifier would read it. A separate Critic then runs with a human verifier's view: the question, the retrieved passages, and the final draft, but not the Drafter's reasoning. It audits each claim against its cited source, so the output is an answer paired with a per-claim audit trace. We also score correctness \emph{strictly}: an answer counts as correct only if it reaches the right disposition under an LLM-as-judge rubric and every citation it emits resolves to a passage the retriever returned.

We evaluate \gandr against five baselines, from zero-shot prompting to advanced RAG and multi-agent self-correction, all sharing one open-source backbone~\citep{nvidia2026nemotroncascade2}, one retrieval surface, and the same citation instruction, so differences reflect architecture rather than model scale or retrieval luck. We contribute (i)~a two-agent system that emits per-claim verification as a generation artifact, a support signal no baseline emits in its standard configuration; (ii)~an evaluation protocol that scores grounding deterministically rather than in aggregate and applies it uniformly across systems and repeated runs as Stable~\&~Correct~@~$k$; and (iii)~evidence, on a $185$-item benchmark from LegalBench~\citep{guha2023legalbench} and LegalBench-RAG~\citep{pipitone2024legalbenchrag}, that the lead is significant against every baseline on the shared backbone and remains positive across four backbones.

\section{Related Work}
\label{sec:related}

\paragraph{Faithfulness and per-claim attribution.}
Since the original RAG architecture~\citep{lewis2020retrieval,huang2023hallucinationsurvey}, a growing line documents answers whose disposition is correct but whose citations are unreliable or post-rationalized~\citep{gao2023alce,niu2024ragtruth,liu2023ragfaithfulness,li2024citationhallucination,wallat2025correctness}. Atomic decomposition is established for \emph{post-hoc} evaluation: FActScore~\citep{min2023factscore} and AttributionBench~\citep{li2024attributionbench} score text claim by claim after the fact, and in law CitaLaw~\citep{zhang2025citalaw}, CourtReasoner~\citep{han2025courtreasoner}, and PRBench~\citep{akyurek2026prbench} grade per-citation support to \emph{measure} frontier models. Structured legal prompting~\citep{yu2022legalprompting} uses IRAC/CREAC schemas to raise task accuracy. What distinguishes \gandr is not a new verification algorithm but how one is placed in the system: it ships the per-claim audit as a first-class artifact with every answer, gates the commit on a deterministic structural check rather than the auditor's own verdict, and exits fail-closed when a draft cannot pass. The atomic check is the post-hoc method of this line, carried inside a system whose commit rule does not trust it.

\paragraph{Retrieval quality as the lever.}
A second family attacks grounding from the retrieval side, improving what reaches the context window through dense retrieval and cross-encoder re-ranking~\citep{karpukhin2020dpr,nogueira2019passagererank}. Better evidence is necessary but not sufficient, since a passage can be retrieved correctly yet cited for a claim it does not support. We therefore include an advanced-RAG baseline that varies retrieval quality while holding generation fixed.

\paragraph{Self-correction and judging.}
A broad family of self-correction methods runs the reviewer in the \emph{same} context as the generator~\citep{yao2023react,shinn2023reflexion,madaan2023selfrefine,dhuliawala2024chain,gou2024critic}. \gandr rejects this, keeping the Critic in a separate context so it cannot ratify the reasoning it is meant to audit. We include same-context self-correction baselines built with CrewAI~\citep{crewai2024} and a LangGraph self-correction graph~\citep{langgraph2024}. LLM-as-judge protocols~\citep{zheng2023judging,liu2023geval} carry position and self-preference biases~\citep{wang2023llmevaluators,panickssery2024selfpref}, which we address with a cross-lineage generator/judge split and annotator calibration.

\section{The \gandr System}
\label{sec:method}

\gandr rests on three architectural commitments: a CREAC generation schema that lays every claim out for inspection, a separate-context Critic that audits each claim, and an orchestrator anchored to a structural check rather than the Critic's verdict.

\paragraph{CREAC as a verification contract.}
CREAC (Conclusion, Rule, Explanation, Application, Conclusion), which opens with the answer's conclusion and restates it after the analysis, is the order practitioners scan in when they verify, so we treat it as a contract on the writer: the five blocks appear in order with both conclusions non-empty, each \texttt{\{cite:\,<rule\_id>\}} marker references a retrieved passage by its \texttt{rule\_id} header, and every claim in the explanation and application blocks pairs with a cited rule. A regex \texttt{protocol\_check} enforces block structure with no LLM call, and atomic \texttt{cite\_verify} enforces per-assertion accountability.

\paragraph{Drafter and Critic in disjoint contexts.}
The Drafter is an LLM call against the shared backbone with a CREAC-shaped system prompt and the retrieved top-$8$ passages, and it is not told the Critic's pass criterion. The Critic runs in a \emph{separate} LLM context: fresh KV-cache, distinct system prompt, and a restricted input of the draft, the question, and the retrieved passages, never the Drafter's chain of thought. This information barrier is intended to approximate a human verifier's view. The Critic invokes three tools: \textbf{\texttt{protocol\_check}} (regex-only); atomic \textbf{\texttt{cite\_verify}}, which splits \texttt{[EXPLANATION]}/\texttt{[APPLICATION]} into $(\text{claim},\text{rule\_id},\text{span})$ tuples and verdicts each \texttt{SUPPORTED}/\texttt{PARTIAL}/\texttt{UNSUPPORTED}/\texttt{MISCITED}; and \textbf{\texttt{sef\_rubric\_score}}, six presentation dimensions from the Structured Explanation Framework (SEF)~\citep{qian2026thinkingjustifyingaligninghighstakes} that only phrase rewrite directives and feed no reported number. A single-context loop would give the model both the reasoning that produced the draft and the obligation to critique it, inviting \emph{anchoring} and \emph{rubric-stamping}, which separate contexts prevent. Sharing backbone weights, this barrier blocks the Drafter's chain of thought but not weight-level self-preference. A different-lineage Critic and Drafter-reasoning injection are left to future work. The verbatim Drafter and Critic prompts are reproduced in Appendix~\ref{app:gandr-prompts}, and the call-by-call pipeline in Appendix~\ref{app:pipelines}.

\paragraph{Protocol-anchored orchestrator.}
The Critic emits verdicts and rewrite directives, but the commit rule is anchored to the structural \texttt{protocol\_check}, not the Critic's PASS/REWRITE verdict. If \texttt{protocol\_check} passes on Round~1 the orchestrator commits, and the Critic acts purely as an \emph{auditor}, still writing verdicts to the trace. That is the normal case: $98.6\,\%$ of runs commit on Round~1 and none is rewritten into a passing answer, so the Critic is an auditor first and a rewrite trigger second. On failure its directives are prepended to the Drafter's next-round input for up to $T{=}3$ rounds, and on exhaustion the system returns \texttt{passed=False} with the full trace, as happens on the remaining $13$ runs. The rewrite path therefore earns its place by failing closed, not by improving answers. The threshold in that path is set to $0.50$, fitted to this corpus rather than a held-out split. No headline metric depends on it, since the strict column never consults it and most items commit before the rewrite path is reached. \S\ref{sec:results} reverts it to a $0.95$ gate to test whether the anchor is load-bearing.

\section{Experimental Setup}
\label{sec:setup}

\paragraph{Benchmark and retrieval.}
A $185$-item benchmark drawn from LegalBench~\citep{guha2023legalbench} (15 task tags) and LegalBench-RAG~\citep{pipitone2024legalbenchrag} (contract-NLI, CUAD), stratified across eight domain buckets spanning constitutional law through tort, at $20$--$30$ items each. The retrieval surface is a single BM25 index of $13{,}090$ passages built once from the inline LegalBench contexts and the LegalBench-RAG mirror, and every system receives the same top-$8$, so stability measures \emph{system} behavior rather than retrieval randomness. BM25 acts here mainly as standardized context delivery. Construction details are in Appendix~\ref{app:dataset}.

\paragraph{Systems, backbone, and judge.}
Six systems share one backbone and retrieval surface: \textbf{B1} zero-shot; \textbf{B2} a single LLM given the same CREAC schema and an SEF-style self-check rubric, the strongest baseline; \textbf{B3} advanced RAG (BM25 + dense + cross-encoder rerank); \textbf{B4} a CrewAI Researcher+Writer crew; \textbf{B5} a LangGraph same-context self-correction graph; \textbf{B6} is \gandr. B4 and B5 use their frameworks' default recipes, so they measure the community-default multi-agent configuration rather than a tuned ceiling. B2~vs.~B6 is the head-to-head. Every system carries the same verbatim \textsc{grounding requirement} block instructing it to cite only \texttt{rule\_id} headers from the retrieved context and to follow each claim with an inline \texttt{\{cite:\,rule\_id\}} marker (Appendix~\ref{app:baseline-prompts}). Strict scoring counts a malformed or unresolvable marker as a failure by design, since a citation a reader cannot follow to its source leaves the answer's provenance unverifiable however right its holding. All six run on NVIDIA Nemotron-Cascade-2-30B-A3B (FP8, vLLM)~\citep{nvidia2026nemotroncascade2}, and the judge is Claude Opus 4.7~\citep{anthropic2026opus47}, whose separate lineage mitigates self-preference bias. Judge rubrics are reproduced in Appendix~\ref{app:judge-rubrics} and runtime settings in Appendix~\ref{app:models}.

\paragraph{Metrics and evaluation parity.}
Three primary columns plus one diagnostic. \textbf{Lenient}: judge rubric $\ge 4$ on the operative holding. \textbf{Strict}: lenient \emph{and} a deterministic grounding check requiring at least one inline marker and that every extracted \texttt{rule\_id} resolves to a header in the retrieved top-$8$, so one fabricated citation fails it (Appendix~\ref{app:strict}). \textbf{Stable~\&~Correct~@~$k{=}5$}: judge-equivalent across all $10$ run pairs (the judge rates the two runs' holdings equivalent for every one of the $C(5,2){=}10$ seed pairs), strict-correct, and strict-grounded on every run. The strict metric is computed \emph{once, post-hoc, identically for all six systems}, and never triggers a retry. Lenient and strict are reported on seed~$0$, fixed as the representative run before scoring. Only \gandr runs a rewrite loop, and as \S\ref{sec:method} notes it almost never fires. No baseline is re-drafted or scored against \gandr's Critic. All systems use temperature $0.7$ and \texttt{max\_tokens}$\,{=}\,2048$.

\section{Results}
\label{sec:results}

\begin{table}[t]
\small
\centering
\setlength{\tabcolsep}{6pt}
\caption{Headline metrics on the $185$-item benchmark (accuracy in \%). \emph{$\Delta$}: strict$-$lenient. Lenient and strict use the seed-$0$ run, S\&C@5 all five seeds. The \emph{CREAC} column is transparency only, since only B2 and \gandr receive the schema. \gandr's leads over B2 clear paired McNemar at $p<0.01$, every other baseline at $p<10^{-8}$.}
\begin{tabular}{lccccc}
\toprule
\textbf{System} & \textbf{Lenient} & \textbf{Strict} & \textbf{$\Delta$} & \textbf{S\&C@5} & \textbf{CREAC} \\
\midrule
B1 zero-shot       & 57.3 & 38.4            & $-18.9$ & 0.049             & 0 \%    \\
B2 SEF prompt      & 60.0 & 59.5            & $-0.5$  & 0.297             & 100 \% \\
B3 advanced RAG    & 60.5 & 43.2            & $-17.3$ & 0.049             & 0 \%    \\
B4 CrewAI          & 58.4 & 21.6            & $-36.8$ & 0.000             & 0 \%    \\
B5 LangGraph       & 51.4 & 21.6            & $-29.8$ & 0.000             & 0 \%    \\
\textbf{B6 \gandr}  & \textbf{71.9} & \textbf{70.8} & $\boldsymbol{-1.1}$ & \textbf{0.460} & 100 \% \\
\bottomrule
\end{tabular}
\label{tab:headline}
\end{table}

\paragraph{Headline.}
\gandr leads on every primary column of Table~\ref{tab:headline}, and the cohort separates into three regimes. Without a structured conclusion, free-form generation drifts: B1 and B3 stay competitive on lenient but collapse on strict, emitting citations the verifier rejects or omitting the marker entirely. CREAC structure absorbs that drift, as B2 collapses the strict-lenient gap to $0.5$\,pp and lifts strict $21.1$\,pp over B1. More agent calls without auditing amplify it: B4 and B5 add LLM calls but no per-claim audit, and strict drops to $21.6\,\%$ on both, each extra call another chance to introduce an unresolvable citation. \gandr runs B2's CREAC Drafter, adds the three commitments of \S\ref{sec:method}, and lifts strict $11.3$\,pp over B2. Because $98.6\,\%$ of items commit at Round~1 (Appendix~\ref{app:loop-depth}), the lead reflects the Drafter configuration and the protocol-anchored commit, not rewriting. That configuration drops B2's same-context self-check rubric, so part of the gain may be the prompt change, and a control isolating the rubric (B2 without it) is left to future work. A taxonomy of how each baseline fails the grounding check is given in Appendix~\ref{app:failure}, and a marker-normalization control that forgives citation form in Appendix~\ref{app:per-domain}.

\paragraph{Backbone transfer.}
Changing only the generator, the strict lead over B2 stays positive on three further backbones, from $+6.5$\,pp on GLM-4.7-Flash~\citep{zai2026glm47flash} to $+3.7$ and $+3.2$ on the \texttt{gpt-5.4} pair~\citep{openai2026gpt54} (Table~\ref{tab:configurability}). These are single-run point estimates, so we read them as directional. On GLM-4.7-Flash \gandr leads on strict but trails B2 on S\&C@5 ($0.297$ vs $0.411$), the one stability reversal across panes, which we flag as unexplained. The lead narrows as stronger backbones comply with the citation contract unprompted, but the per-claim trace and the fail-closed guarantee remain architectural properties no backbone supplies on its own.

\begin{table}[t]
\small
\centering
\setlength{\tabcolsep}{6pt}
\caption{Backbone-pluggability check. Strict in \%, S\&C@5 in $[0,1]$, $\Delta$ the \gandr$-$B2 strict point estimate. Only the generator changes across panes. Nemotron values match Table~\ref{tab:headline}.}
\begin{tabular}{llcccc}
\toprule
\textbf{Eco} & \textbf{Backbone} & \textbf{Sys} & \textbf{Strict} & \textbf{S\&C} & \textbf{$\Delta$} \\
\midrule
\multirow{4}{*}{open}   & Nemotron    & B2  & 59.5 & 0.297 & ---   \\
                        & Nemotron    & \gandr & \textbf{70.8} & \textbf{0.460} & \textbf{+11.3} \\
                        & GLM-4.7-Flash & B2 & 62.7 & 0.411 & ---   \\
                        & GLM-4.7-Flash & \gandr & \textbf{69.2} & 0.297 & \textbf{+6.5} \\
\midrule
\multirow{4}{*}{closed} & \texttt{gpt-5.4-mini} & B2 & 74.1 & 0.638 & --- \\
                        & \texttt{gpt-5.4-mini} & \gandr & \textbf{77.8} & \textbf{0.686} & \textbf{+3.7} \\
                        & \texttt{gpt-5.4} & B2 & 76.8 & 0.692 & --- \\
                        & \texttt{gpt-5.4} & \gandr & \textbf{80.0} & \textbf{0.762} & \textbf{+3.2} \\
\bottomrule
\end{tabular}
\label{tab:configurability}
\end{table}

\paragraph{Ablating the commit rule.}
The commit rule is the one architectural choice we can isolate cleanly: reverting it to the original $0.95$ atomic-verification gate drops strict accuracy to $48.1\,\%$, a loss of $22.7$\,pp (Table~\ref{tab:ablations}). That gate exceeds the verifier's reliable signal here, so rewrites fire on almost every item and quality degrades across rounds. The reverted variant returns \texttt{passed=False} on $92.7\,\%$ of runs, but strict still scores each final answer post-hoc, so the $48.1\,\%$ reflects degraded rewrites, not counted refusals. The Critic and the decomposition mode cannot be isolated this way: since the anchor commits without consulting the Critic, disabling either leaves the committed answer unchanged on the $97$--$98\,\%$ of Round-1 commits, so we report neither as an accuracy effect (Appendix~\ref{app:ablation-identifiability}). No result here credits the audit machinery with an accuracy gain. Its demonstrated contribution is the per-claim trace and the fail-closed exit.

\begin{table}[t]
\small
\centering
\setlength{\tabcolsep}{6pt}
\caption{Ablation of the protocol-anchored commit rule (accuracy in \%). The variant reverts the commit gate to the original $0.95$ atomic-verification threshold, holding the Drafter, benchmark, retrieval, and judge fixed. See Appendix~\ref{app:ablation-identifiability} for why the Critic and \texttt{cite\_verify} variants are not separately identifiable.}
\begin{tabular}{lccc}
\toprule
\textbf{Variant} & \textbf{Lenient} & \textbf{Strict} & \textbf{$\Delta$~Strict} \\
\midrule
\textbf{\gandr (full)}    & \textbf{71.9} & \textbf{70.8} & ---           \\
$-$ Round-1 anchor       & 49.2          & 48.1          & $-22.7$    \\
\bottomrule
\end{tabular}
\label{tab:ablations}
\end{table}

\begin{figure}[t]
\centering
\includegraphics[width=0.95\linewidth]{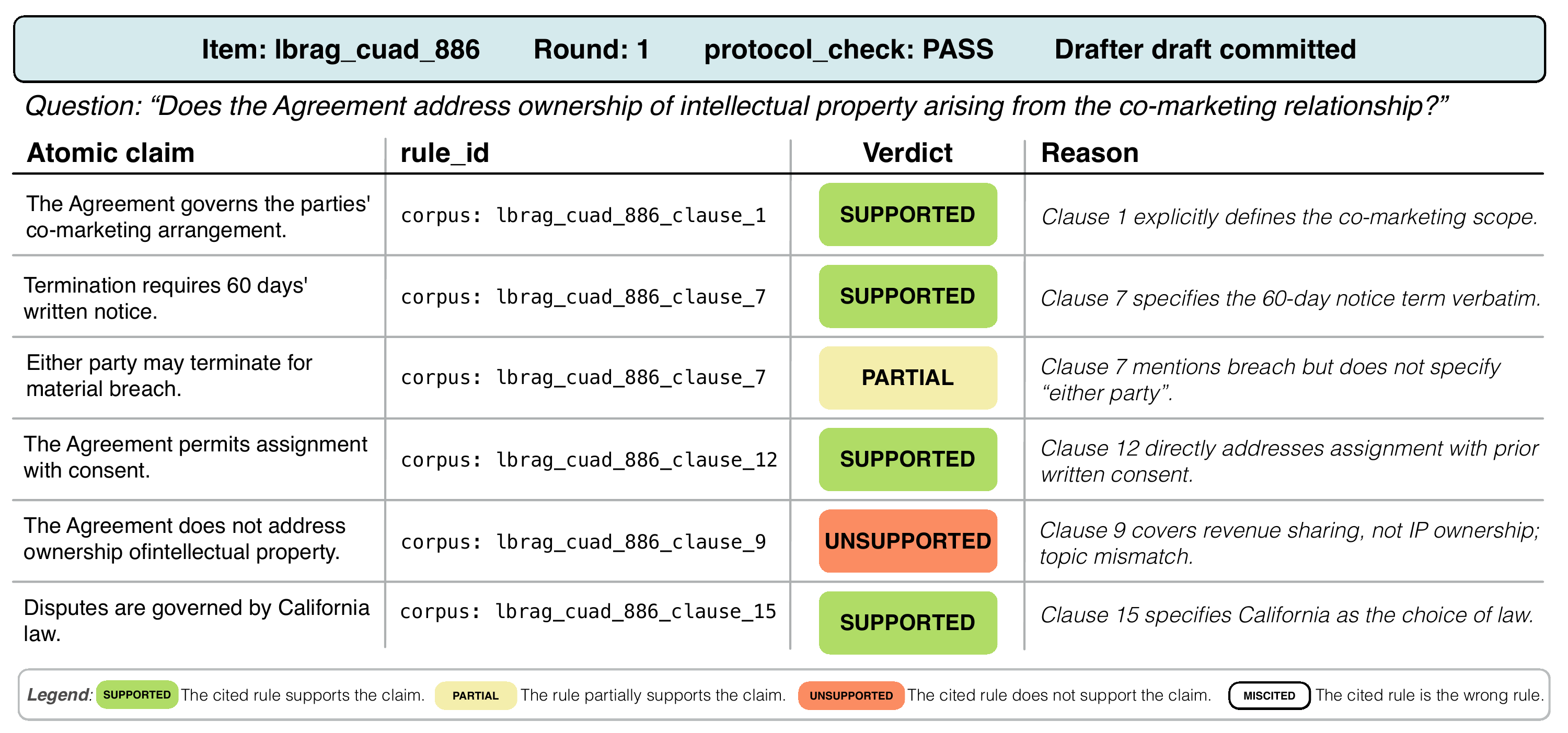}
\caption{Atomic-claim audit trace from \gandr on \texttt{lbrag\_cuad\_2886}: six per-claim verdicts. The \texttt{UNSUPPORTED} verdict on claim~5 catches a topic mismatch, and \texttt{protocol\_check} still passes, so the trace surfaces the under-supported claim rather than blocking commit.}
\label{fig:audit-trace}
\end{figure}

\paragraph{Auditability.}
Alongside each answer \gandr ships an atomic-claim audit trace, one claim--citation--verdict triple per assertion (Figure~\ref{fig:audit-trace}). It flags citations that resolve to a retrieved passage yet only topically match the claim, an orthogonal signal strict accuracy cannot see. Against two law-trained annotators on $148$ blinded assertions it flags such claims at precision $0.80$ and recall $0.88$, though its four-way labels are advisory (Appendix~\ref{app:cite-verify-validation}).

\paragraph{Significance and cost.}
Per-domain buckets ($20$--$30$ items) are too small to separate systems individually, so the claim rests on the aggregate: among the items the two systems disagree on, \gandr is correct on $38$ and B2 on $17$ (paired McNemar $p{=}0.006$; per-domain intervals and all pairwise tests in Appendix~\ref{app:per-domain}). The judge tracks two law-trained annotators closely (Pearson $r{=}0.846$, quadratic-weighted $\kappa_q{=}0.845$; Appendix~\ref{app:calibration}). Because the Round-1 anchor commits most items in one round, metered cost is $3.0$--$3.3\times$ B2 rather than the $10\times$ its call count implies (Appendix~\ref{app:cost}).

\section{Conclusion}
\label{sec:conclusion}

Legal answers need verification claim by claim, not a score in aggregate. \gandr builds that verification into generation, leading the strongest baseline by $11.3$ strict points on the shared backbone ($p<0.01$), staying positive at $+3.2$ to $+6.5$ points on three further backbones, and shipping a per-claim audit trace, validated at F1 $0.84$, with every answer it does not refuse. The benchmark is classification- and NLI-weighted and indexes each item's own context, so \gandr is validated as a citation-discipline layer over supplied authority, not open-corpus legal research. Multi-source memos and harder retrieval are left to future work.

\bibliographystyle{plainnat}
\bibliography{references}

\begin{thebibliography}{35}
\providecommand{\natexlab}[1]{#1}
\providecommand{\url}[1]{\texttt{#1}}
\expandafter\ifx\csname urlstyle\endcsname\relax
  \providecommand{\doi}[1]{doi: #1}\else
  \providecommand{\doi}{doi: \begingroup \urlstyle{rm}\Url}\fi

\bibitem[Aky{\"u}rek et~al.(2026)Aky{\"u}rek, Gosai, Zhang, Gupta, Jeong,
  Gunjal, Rabbani, Mazzone, IV, Meymand, Chattha, Rodriguez, Buendia, Singh,
  Liu, Chawla, Cline, Ogaz, Montoya, Wang, Bhatter, Ayestaran, Liu, and
  He]{akyurek2026prbench}
Afra~Feyza Aky{\"u}rek, Advait Gosai, Chen Bo~Calvin Zhang, Vipul Gupta,
  Jaehwan Jeong, Anisha Gunjal, Tahseen Rabbani, Maria Mazzone, David~Randolph
  IV, Mohammad~Mahmoudi Meymand, Gurshaan Chattha, Paula Rodriguez, Diego
  A.~Mares Buendia, Pavit Singh, Michael Liu, Subodh Chawla, Peter Cline, Lucy
  Ogaz, Ernesto Gabriel~Hern{\'a}ndez Montoya, Zihao Wang, Pavi Bhatter, Marcos
  Ayestaran, Bing Liu, and Yunzhong He.
\newblock {PRB}ench: Large-scale expert rubrics for evaluating high-stakes
  professional reasoning.
\newblock In Maria Liakata, Viviane~P. Moreira, Jiajun Zhang, and David
  Jurgens, editors, \emph{Proceedings of the 64th Annual Meeting of the
  {A}ssociation for {C}omputational {L}inguistics (Volume 1: Long Papers)},
  pages 42297--42325, San Diego, California, United States, July 2026.
  Association for Computational Linguistics.
\newblock ISBN 979-8-89176-390-6.
\newblock \doi{10.18653/v1/2026.acl-long.1958}.
\newblock URL \url{https://aclanthology.org/2026.acl-long.1958/}.

\bibitem[{Anthropic}(2026)]{anthropic2026opus47}
{Anthropic}.
\newblock Introducing claude opus 4.7.
\newblock \url{https://www.anthropic.com/news/claude-opus-4-7}, may 2026.
\newblock Retrieved: May 22, 2026.

\bibitem[CrewAI(2026)]{crewai2024}
CrewAI.
\newblock The leading multi-agent platform.
\newblock \url{https://crewai.com}, may 2026.
\newblock Retrieved: May 22, 2026.

\bibitem[Dahl et~al.(2024)Dahl, Magesh, Suzgun, and Ho]{dahl2024legalfictions}
Matthew Dahl, Varun Magesh, Mirac Suzgun, and Daniel~E Ho.
\newblock Large legal fictions: Profiling legal hallucinations in large
  language models.
\newblock \emph{Journal of Legal Analysis}, 16\penalty0 (1):\penalty0 64--93,
  01 2024.
\newblock ISSN 2161-7201.
\newblock \doi{10.1093/jla/laae003}.
\newblock URL \url{https://doi.org/10.1093/jla/laae003}.

\bibitem[Dhuliawala et~al.(2024)Dhuliawala, Komeili, Xu, Raileanu, Li,
  Celikyilmaz, and Weston]{dhuliawala2024chain}
Shehzaad Dhuliawala, Mojtaba Komeili, Jing Xu, Roberta Raileanu, Xian Li, Asli
  Celikyilmaz, and Jason Weston.
\newblock Chain-of-verification reduces hallucination in large language models.
\newblock In Lun-Wei Ku, Andre Martins, and Vivek Srikumar, editors,
  \emph{Findings of the Association for Computational Linguistics: ACL 2024},
  pages 3563--3578, Bangkok, Thailand, August 2024. Association for
  Computational Linguistics.
\newblock \doi{10.18653/v1/2024.findings-acl.212}.
\newblock URL \url{https://aclanthology.org/2024.findings-acl.212/}.

\bibitem[Gao et~al.(2023)Gao, Yen, Yu, and Chen]{gao2023alce}
Tianyu Gao, Howard Yen, Jiatong Yu, and Danqi Chen.
\newblock Enabling large language models to generate text with citations.
\newblock In Houda Bouamor, Juan Pino, and Kalika Bali, editors,
  \emph{Proceedings of the 2023 Conference on Empirical Methods in Natural
  Language Processing}, pages 6465--6488, Singapore, December 2023. Association
  for Computational Linguistics.
\newblock \doi{10.18653/v1/2023.emnlp-main.398}.
\newblock URL \url{https://aclanthology.org/2023.emnlp-main.398/}.

\bibitem[Gou et~al.(2024)Gou, Shao, Gong, Shen, Yang, Duan, and
  Chen]{gou2024critic}
Zhibin Gou, Zhihong Shao, Yeyun Gong, Yelong Shen, Yujiu Yang, Nan Duan, and
  Weizhu Chen.
\newblock Critic: Large language models can self-correct with tool-interactive
  critiquing, 2024.
\newblock URL \url{https://arxiv.org/abs/2305.11738}.

\bibitem[Guha et~al.(2023)Guha, Nyarko, Ho, R\'{e}, Chilton, Narayana,
  Chohlas-Wood, Peters, Waldon, Rockmore, Zambrano, Talisman, Hoque, Surani,
  Fagan, Sarfaty, Dickinson, Porat, Hegland, Wu, Nudell, Niklaus, Nay, Choi,
  Tobia, Hagan, Ma, Livermore, Rasumov-Rahe, Holzenberger, Kolt, Henderson,
  Rehaag, Goel, Gao, Williams, Gandhi, Zur, Iyer, and Li]{guha2023legalbench}
Neel Guha, Julian Nyarko, Daniel~E. Ho, Christopher R\'{e}, Adam Chilton,
  Aditya Narayana, Alex Chohlas-Wood, Austin Peters, Brandon Waldon, Daniel~N.
  Rockmore, Diego Zambrano, Dmitry Talisman, Enam Hoque, Faiz Surani, Frank
  Fagan, Galit Sarfaty, Gregory~M. Dickinson, Haggai Porat, Jason Hegland,
  Jessica Wu, Joe Nudell, Joel Niklaus, John Nay, Jonathan~H. Choi, Kevin
  Tobia, Margaret Hagan, Megan Ma, Michael Livermore, Nikon Rasumov-Rahe, Nils
  Holzenberger, Noam Kolt, Peter Henderson, Sean Rehaag, Sharad Goel, Shang
  Gao, Spencer Williams, Sunny Gandhi, Tom Zur, Varun Iyer, and Zehua Li.
\newblock Legalbench: a collaboratively built benchmark for measuring legal
  reasoning in large language models.
\newblock In \emph{Proceedings of the 37th International Conference on Neural
  Information Processing Systems}, NIPS '23, Red Hook, NY, USA, 2023. Curran
  Associates Inc.
\newblock URL \url{https://dl.acm.org/doi/10.5555/3666122.3668037}.

\bibitem[Han et~al.(2025)Han, Takashima, Shen, Liu, Liu, Thuo, Knowlton,
  Piskac, Shapiro, and Cohan]{han2025courtreasoner}
Sophia~Simeng Han, Yoshiki Takashima, Shannon~Zejiang Shen, Chen Liu, Yixin
  Liu, Roque~K. Thuo, Sonia Knowlton, Ruzica Piskac, Scott~J Shapiro, and Arman
  Cohan.
\newblock {C}ourt{R}easoner: Can {LLM} agents reason like judges?
\newblock In Christos Christodoulopoulos, Tanmoy Chakraborty, Carolyn Rose, and
  Violet Peng, editors, \emph{Proceedings of the 2025 Conference on Empirical
  Methods in Natural Language Processing}, pages 35291--35306, Suzhou, China,
  November 2025. Association for Computational Linguistics.
\newblock ISBN 979-8-89176-332-6.
\newblock \doi{10.18653/v1/2025.emnlp-main.1787}.
\newblock URL \url{https://aclanthology.org/2025.emnlp-main.1787/}.

\bibitem[Huang et~al.(2025)Huang, Yu, Ma, Zhong, Feng, Wang, Chen, Peng, Feng,
  Qin, and Liu]{huang2023hallucinationsurvey}
Lei Huang, Weijiang Yu, Weitao Ma, Weihong Zhong, Zhangyin Feng, Haotian Wang,
  Qianglong Chen, Weihua Peng, Xiaocheng Feng, Bing Qin, and Ting Liu.
\newblock A survey on hallucination in large language models: Principles,
  taxonomy, challenges, and open questions.
\newblock \emph{ACM Trans. Inf. Syst.}, 43\penalty0 (2), January 2025.
\newblock ISSN 1046-8188.
\newblock \doi{10.1145/3703155}.
\newblock URL \url{https://doi.org/10.1145/3703155}.

\bibitem[Karpukhin et~al.(2020)Karpukhin, Oguz, Min, Lewis, Wu, Edunov, Chen,
  and Yih]{karpukhin2020dpr}
Vladimir Karpukhin, Barlas Oguz, Sewon Min, Patrick Lewis, Ledell Wu, Sergey
  Edunov, Danqi Chen, and Wen-tau Yih.
\newblock Dense passage retrieval for open-domain question answering.
\newblock In Bonnie Webber, Trevor Cohn, Yulan He, and Yang Liu, editors,
  \emph{Proceedings of the 2020 Conference on Empirical Methods in Natural
  Language Processing (EMNLP)}, pages 6769--6781, Online, November 2020.
  Association for Computational Linguistics.
\newblock \doi{10.18653/v1/2020.emnlp-main.550}.
\newblock URL \url{https://aclanthology.org/2020.emnlp-main.550/}.

\bibitem[Langchain(2026)]{langgraph2024}
Langchain.
\newblock Langgraph: Agent orchestration framework for reliable ai agents.
\newblock \url{https://www.langchain.com/langgraph}, may 2026.
\newblock Retrieved: May 22, 2026.

\bibitem[Lewis et~al.(2020)Lewis, Perez, Piktus, Petroni, Karpukhin, Goyal,
  K\"{u}ttler, Lewis, Yih, Rockt\"{a}schel, Riedel, and
  Kiela]{lewis2020retrieval}
Patrick Lewis, Ethan Perez, Aleksandra Piktus, Fabio Petroni, Vladimir
  Karpukhin, Naman Goyal, Heinrich K\"{u}ttler, Mike Lewis, Wen-tau Yih, Tim
  Rockt\"{a}schel, Sebastian Riedel, and Douwe Kiela.
\newblock Retrieval-augmented generation for knowledge-intensive nlp tasks.
\newblock In \emph{Proceedings of the 34th International Conference on Neural
  Information Processing Systems}, NIPS '20, Red Hook, NY, USA, 2020. Curran
  Associates Inc.
\newblock ISBN 9781713829546.
\newblock URL \url{https://dl.acm.org/doi/abs/10.5555/3495724.3496517}.

\bibitem[Li et~al.(2024{\natexlab{a}})Li, Cao, Pan, Ma, and
  Sun]{li2024citationhallucination}
Xinze Li, Yixin Cao, Liangming Pan, Yubo Ma, and Aixin Sun.
\newblock Towards verifiable generation: A benchmark for knowledge-aware
  language model attribution.
\newblock In Lun-Wei Ku, Andre Martins, and Vivek Srikumar, editors,
  \emph{Findings of the Association for Computational Linguistics: ACL 2024},
  pages 493--516, Bangkok, Thailand, August 2024{\natexlab{a}}. Association for
  Computational Linguistics.
\newblock \doi{10.18653/v1/2024.findings-acl.28}.
\newblock URL \url{https://aclanthology.org/2024.findings-acl.28/}.

\bibitem[Li et~al.(2024{\natexlab{b}})Li, Yue, Liao, and
  Sun]{li2024attributionbench}
Yifei Li, Xiang Yue, Zeyi Liao, and Huan Sun.
\newblock {A}ttribution{B}ench: How hard is automatic attribution evaluation?
\newblock In Lun-Wei Ku, Andre Martins, and Vivek Srikumar, editors,
  \emph{Findings of the Association for Computational Linguistics: ACL 2024},
  pages 14919--14935, Bangkok, Thailand, August 2024{\natexlab{b}}. Association
  for Computational Linguistics.
\newblock \doi{10.18653/v1/2024.findings-acl.886}.
\newblock URL \url{https://aclanthology.org/2024.findings-acl.886/}.

\bibitem[Liu et~al.(2023{\natexlab{a}})Liu, Zhang, and
  Liang]{liu2023ragfaithfulness}
Nelson Liu, Tianyi Zhang, and Percy Liang.
\newblock Evaluating verifiability in generative search engines.
\newblock In Houda Bouamor, Juan Pino, and Kalika Bali, editors, \emph{Findings
  of the Association for Computational Linguistics: EMNLP 2023}, pages
  7001--7025, Singapore, December 2023{\natexlab{a}}. Association for
  Computational Linguistics.
\newblock \doi{10.18653/v1/2023.findings-emnlp.467}.
\newblock URL \url{https://aclanthology.org/2023.findings-emnlp.467/}.

\bibitem[Liu et~al.(2023{\natexlab{b}})Liu, Iter, Xu, Wang, Xu, and
  Zhu]{liu2023geval}
Yang Liu, Dan Iter, Yichong Xu, Shuohang Wang, Ruochen Xu, and Chenguang Zhu.
\newblock {G}-eval: {NLG} evaluation using gpt-4 with better human alignment.
\newblock In Houda Bouamor, Juan Pino, and Kalika Bali, editors,
  \emph{Proceedings of the 2023 Conference on Empirical Methods in Natural
  Language Processing}, pages 2511--2522, Singapore, December
  2023{\natexlab{b}}. Association for Computational Linguistics.
\newblock \doi{10.18653/v1/2023.emnlp-main.153}.
\newblock URL \url{https://aclanthology.org/2023.emnlp-main.153/}.

\bibitem[Madaan et~al.(2023)Madaan, Tandon, Gupta, Hallinan, Gao, Wiegreffe,
  Alon, Dziri, Prabhumoye, Yang, Gupta, Majumder, Hermann, Welleck,
  Yazdanbakhsh, and Clark]{madaan2023selfrefine}
Aman Madaan, Niket Tandon, Prakhar Gupta, Skyler Hallinan, Luyu Gao, Sarah
  Wiegreffe, Uri Alon, Nouha Dziri, Shrimai Prabhumoye, Yiming Yang, Shashank
  Gupta, Bodhisattwa~Prasad Majumder, Katherine Hermann, Sean Welleck, Amir
  Yazdanbakhsh, and Peter Clark.
\newblock Self-refine: iterative refinement with self-feedback.
\newblock In \emph{Proceedings of the 37th International Conference on Neural
  Information Processing Systems}, NIPS '23, Red Hook, NY, USA, 2023. Curran
  Associates Inc.
\newblock URL \url{https://dl.acm.org/doi/10.5555/3666122.3668141}.

\bibitem[Magesh et~al.(2025)Magesh, Surani, Dahl, Suzgun, Manning, and
  Ho]{magesh2024hallucinationfree}
Varun Magesh, Faiz Surani, Matthew Dahl, Mirac Suzgun, Christopher~D. Manning,
  and Daniel~E. Ho.
\newblock Hallucination-free? assessing the reliability of leading ai legal
  research tools.
\newblock \emph{Journal of Empirical Legal Studies}, 22\penalty0 (2):\penalty0
  216--242, 2025.
\newblock \doi{https://doi.org/10.1111/jels.12413}.
\newblock URL \url{https://onlinelibrary.wiley.com/doi/abs/10.1111/jels.12413}.

\bibitem[Min et~al.(2023)Min, Krishna, Lyu, Lewis, Yih, Koh, Iyyer,
  Zettlemoyer, and Hajishirzi]{min2023factscore}
Sewon Min, Kalpesh Krishna, Xinxi Lyu, Mike Lewis, Wen-tau Yih, Pang Koh, Mohit
  Iyyer, Luke Zettlemoyer, and Hannaneh Hajishirzi.
\newblock {FA}ct{S}core: Fine-grained atomic evaluation of factual precision in
  long form text generation.
\newblock In Houda Bouamor, Juan Pino, and Kalika Bali, editors,
  \emph{Proceedings of the 2023 Conference on Empirical Methods in Natural
  Language Processing}, pages 12076--12100, Singapore, December 2023.
  Association for Computational Linguistics.
\newblock \doi{10.18653/v1/2023.emnlp-main.741}.
\newblock URL \url{https://aclanthology.org/2023.emnlp-main.741/}.

\bibitem[Niu et~al.(2024)Niu, Wu, Zhu, Xu, Shum, Zhong, Song, and
  Zhang]{niu2024ragtruth}
Cheng Niu, Yuanhao Wu, Juno Zhu, Siliang Xu, KaShun Shum, Randy Zhong, Juntong
  Song, and Tong Zhang.
\newblock {RAGT}ruth: A hallucination corpus for developing trustworthy
  retrieval-augmented language models.
\newblock In Lun-Wei Ku, Andre Martins, and Vivek Srikumar, editors,
  \emph{Proceedings of the 62nd Annual Meeting of the Association for
  Computational Linguistics (Volume 1: Long Papers)}, pages 10862--10878,
  Bangkok, Thailand, August 2024. Association for Computational Linguistics.
\newblock \doi{10.18653/v1/2024.acl-long.585}.
\newblock URL \url{https://aclanthology.org/2024.acl-long.585/}.

\bibitem[Nogueira and Cho(2020)]{nogueira2019passagererank}
Rodrigo Nogueira and Kyunghyun Cho.
\newblock Passage re-ranking with bert, 2020.
\newblock URL \url{https://arxiv.org/abs/1901.04085}.

\bibitem[{OpenAI}(2026)]{openai2026gpt54}
{OpenAI}.
\newblock Introducing gpt‑5.4 mini and nano.
\newblock \url{https://openai.com/index/introducing-gpt-5-4-mini-and-nano/},
  may 2026.
\newblock Retrieved: May 22, 2026.

\bibitem[Panickssery et~al.(2024)Panickssery, Bowman, and
  Feng]{panickssery2024selfpref}
Arjun Panickssery, Samuel~R. Bowman, and Shi Feng.
\newblock Llm evaluators recognize and favor their own generations.
\newblock In \emph{Proceedings of the 38th International Conference on Neural
  Information Processing Systems}, NIPS '24, Red Hook, NY, USA, 2024. Curran
  Associates Inc.
\newblock ISBN 9798331314385.
\newblock URL \url{https://dl.acm.org/doi/10.5555/3737916.3740113}.

\bibitem[Pipitone and Alami(2024)]{pipitone2024legalbenchrag}
Nicholas Pipitone and Ghita~Houir Alami.
\newblock Legalbench-rag: A benchmark for retrieval-augmented generation in the
  legal domain, 2024.
\newblock URL \url{https://arxiv.org/abs/2408.10343}.

\bibitem[Qian et~al.(2026)Qian, Wang, Chen, Wu, and
  Stathopoulos]{qian2026thinkingjustifyingaligninghighstakes}
Chen Qian, Yimeng Wang, Yu~Chen, Lingfei Wu, and Andreas Stathopoulos.
\newblock From ``thinking'' to ``justifying'': Aligning high-stakes
  explainability with professional communication standards.
\newblock In Maria Liakata, Viviane~P. Moreira, Jiajun Zhang, and David
  Jurgens, editors, \emph{Findings of the {A}ssociation for {C}omputational
  {L}inguistics: {ACL} 2026}, pages 24628--24637, San Diego, California, United
  States, July 2026. Association for Computational Linguistics.
\newblock ISBN 979-8-89176-395-1.
\newblock \doi{10.18653/v1/2026.findings-acl.1232}.
\newblock URL \url{https://aclanthology.org/2026.findings-acl.1232/}.

\bibitem[Shinn et~al.(2023)Shinn, Cassano, Gopinath, Narasimhan, and
  Yao]{shinn2023reflexion}
Noah Shinn, Federico Cassano, Ashwin Gopinath, Karthik Narasimhan, and Shunyu
  Yao.
\newblock Reflexion: language agents with verbal reinforcement learning.
\newblock In \emph{Proceedings of the 37th International Conference on Neural
  Information Processing Systems}, NIPS '23, Red Hook, NY, USA, 2023. Curran
  Associates Inc.
\newblock URL \url{https://dl.acm.org/doi/10.5555/3666122.3666499}.

\bibitem[Team et~al.(2025)Team, Zeng, Lv, Zheng, Hou, Chen, Xie, Wang, Yin,
  Zeng, Zhang, Wang, Zhong, Liu, Lu, Cao, Zhang, Huang, Wei, Cheng, An, Niu,
  Wen, Bai, Du, Wang, Zhu, Zhang, Wen, Wu, Xu, Huang, Zhao, Cai, Yu, Li, Ge,
  Huang, Zhang, Xu, Zhu, Li, Yin, Lin, Yang, Jiang, Ai, Zhu, Wang, Pan, Wang,
  Sun, Li, Li, Hu, Zhang, Peng, Tai, Zhang, Wang, Yang, Liu, Zhao, Liu, Yan,
  Liu, Chen, Li, Zhao, Ren, Jiao, Zhao, Yan, Wang, Gui, Zhao, Liu, Li, Li, Lu,
  Wang, Yuan, Li, Du, Du, Liu, Zhi, Gao, Wang, Yang, Xu, Fan, Wu, Ding, Wang,
  Zhang, Li, Xu, Zhao, Zhai, Du, Dong, Lei, Tu, Yang, Lu, Li, Li, Shuang-Li,
  Yang, Yi, Yu, Tian, Wang, Yu, Tam, Liang, Liu, Wang, Jia, Gu, Ling, Wang,
  Fan, Pan, Zhang, Zhang, Fu, Zhang, Xu, Wu, Lu, Wang, Zhou, Pan, Zhang, Wang,
  Li, Su, Geng, Zhu, Yang, Li, Wu, Li, Liu, Wang, Li, Zhang, Liu, Yang, Zhou,
  Qiao, Feng, Liu, Zhang, Wang, Yao, Wang, Liu, Chai, Li, Zhao, Chen, Zhai, Xu,
  Huang, Wang, Li, Dong, and Tang]{zai2026glm47flash}
GLM Team, Aohan Zeng, Xin Lv, Qinkai Zheng, Zhenyu Hou, Bin Chen, Chengxing
  Xie, Cunxiang Wang, Da~Yin, Hao Zeng, Jiajie Zhang, Kedong Wang, Lucen Zhong,
  Mingdao Liu, Rui Lu, Shulin Cao, Xiaohan Zhang, Xuancheng Huang, Yao Wei,
  Yean Cheng, Yifan An, Yilin Niu, Yuanhao Wen, Yushi Bai, Zhengxiao Du, Zihan
  Wang, Zilin Zhu, Bohan Zhang, Bosi Wen, Bowen Wu, Bowen Xu, Can Huang, Casey
  Zhao, Changpeng Cai, Chao Yu, Chen Li, Chendi Ge, Chenghua Huang, Chenhui
  Zhang, Chenxi Xu, Chenzheng Zhu, Chuang Li, Congfeng Yin, Daoyan Lin, Dayong
  Yang, Dazhi Jiang, Ding Ai, Erle Zhu, Fei Wang, Gengzheng Pan, Guo Wang,
  Hailong Sun, Haitao Li, Haiyang Li, Haiyi Hu, Hanyu Zhang, Hao Peng, Hao Tai,
  Haoke Zhang, Haoran Wang, Haoyu Yang, He~Liu, He~Zhao, Hongwei Liu, Hongxi
  Yan, Huan Liu, Huilong Chen, Ji~Li, Jiajing Zhao, Jiamin Ren, Jian Jiao,
  Jiani Zhao, Jianyang Yan, Jiaqi Wang, Jiayi Gui, Jiayue Zhao, Jie Liu, Jijie
  Li, Jing Li, Jing Lu, Jingsen Wang, Jingwei Yuan, Jingxuan Li, Jingzhao Du,
  Jinhua Du, Jinxin Liu, Junkai Zhi, Junli Gao, Ke~Wang, Lekang Yang, Liang Xu,
  Lin Fan, Lindong Wu, Lintao Ding, Lu~Wang, Man Zhang, Minghao Li, Minghuan
  Xu, Mingming Zhao, Mingshu Zhai, Pengfan Du, Qian Dong, Shangde Lei,
  Shangqing Tu, Shangtong Yang, Shaoyou Lu, Shijie Li, Shuang Li, Shuang-Li,
  Shuxun Yang, Sibo Yi, Tianshu Yu, Wei Tian, Weihan Wang, Wenbo Yu, Weng~Lam
  Tam, Wenjie Liang, Wentao Liu, Xiao Wang, Xiaohan Jia, Xiaotao Gu, Xiaoying
  Ling, Xin Wang, Xing Fan, Xingru Pan, Xinyuan Zhang, Xinze Zhang, Xiuqing Fu,
  Xunkai Zhang, Yabo Xu, Yandong Wu, Yida Lu, Yidong Wang, Yilin Zhou, Yiming
  Pan, Ying Zhang, Yingli Wang, Yingru Li, Yinpei Su, Yipeng Geng, Yitong Zhu,
  Yongkun Yang, Yuhang Li, Yuhao Wu, Yujiang Li, Yunan Liu, Yunqing Wang,
  Yuntao Li, Yuxuan Zhang, Zezhen Liu, Zhen Yang, Zhengda Zhou, Zhongpei Qiao,
  Zhuoer Feng, Zhuorui Liu, Zichen Zhang, Zihan Wang, Zijun Yao, Zikang Wang,
  Ziqiang Liu, Ziwei Chai, Zixuan Li, Zuodong Zhao, Wenguang Chen, Jidong Zhai,
  Bin Xu, Minlie Huang, Hongning Wang, Juanzi Li, Yuxiao Dong, and Jie Tang.
\newblock Glm-4.5: Agentic, reasoning, and coding (arc) foundation models,
  2025.
\newblock URL \url{https://arxiv.org/abs/2508.06471}.

\bibitem[Wallat et~al.(2025)Wallat, Heuss, Rijke, and
  Anand]{wallat2025correctness}
Jonas Wallat, Maria Heuss, Maarten~de Rijke, and Avishek Anand.
\newblock Correctness is not faithfulness in retrieval augmented generation
  attributions.
\newblock In \emph{Proceedings of the 2025 International ACM SIGIR Conference
  on Innovative Concepts and Theories in Information Retrieval (ICTIR)}, ICTIR
  '25, page 22–32, New York, NY, USA, 2025. Association for Computing
  Machinery.
\newblock ISBN 9798400718618.
\newblock \doi{10.1145/3731120.3744592}.
\newblock URL \url{https://doi.org/10.1145/3731120.3744592}.

\bibitem[Wang et~al.(2024)Wang, Li, Chen, Cai, Zhu, Lin, Cao, Kong, Liu, Liu,
  and Sui]{wang2023llmevaluators}
Peiyi Wang, Lei Li, Liang Chen, Zefan Cai, Dawei Zhu, Binghuai Lin, Yunbo Cao,
  Lingpeng Kong, Qi~Liu, Tianyu Liu, and Zhifang Sui.
\newblock Large language models are not fair evaluators.
\newblock In Lun-Wei Ku, Andre Martins, and Vivek Srikumar, editors,
  \emph{Proceedings of the 62nd Annual Meeting of the Association for
  Computational Linguistics (Volume 1: Long Papers)}, pages 9440--9450,
  Bangkok, Thailand, August 2024. Association for Computational Linguistics.
\newblock \doi{10.18653/v1/2024.acl-long.511}.
\newblock URL \url{https://aclanthology.org/2024.acl-long.511/}.

\bibitem[Yang et~al.(2026)Yang, Liu, Chen, Dai, Wang, Lin, Lee, Chen, Jiang,
  He, Pi, Lam, Lee, Bukharin, Shoeybi, Catanzaro, and
  Ping]{nvidia2026nemotroncascade2}
Zhuolin Yang, Zihan Liu, Yang Chen, Wenliang Dai, Boxin Wang, Sheng-Chieh Lin,
  Chankyu Lee, Yangyi Chen, Dongfu Jiang, Jiafan He, Renjie Pi, Grace Lam,
  Nayeon Lee, Alexander Bukharin, Mohammad Shoeybi, Bryan Catanzaro, and Wei
  Ping.
\newblock Nemotron-cascade 2: Post-training llms with cascade rl and
  multi-domain on-policy distillation, 2026.
\newblock URL \url{https://arxiv.org/abs/2603.19220}.

\bibitem[Yao et~al.(2023)Yao, Zhao, Yu, Du, Shafran, Narasimhan, and
  Cao]{yao2023react}
Shunyu Yao, Jeffrey Zhao, Dian Yu, Nan Du, Izhak Shafran, Karthik Narasimhan,
  and Yuan Cao.
\newblock React: Synergizing reasoning and acting in language models, 2023.
\newblock URL \url{https://arxiv.org/abs/2210.03629}.

\bibitem[Yu et~al.(2022)Yu, Quartey, and Schilder]{yu2022legalprompting}
Fangyi Yu, Lee Quartey, and Frank Schilder.
\newblock Legal prompting: Teaching a language model to think like a lawyer,
  2022.
\newblock URL \url{https://arxiv.org/abs/2212.01326}.

\bibitem[Zhang et~al.(2025)Zhang, Yu, Dai, and Xu]{zhang2025citalaw}
Kepu Zhang, Weijie Yu, Sunhao Dai, and Jun Xu.
\newblock {C}ita{L}aw: Enhancing {LLM} with citations in legal domain.
\newblock In Wanxiang Che, Joyce Nabende, Ekaterina Shutova, and Mohammad~Taher
  Pilehvar, editors, \emph{Findings of the Association for Computational
  Linguistics: ACL 2025}, pages 11183--11196, Vienna, Austria, July 2025.
  Association for Computational Linguistics.
\newblock ISBN 979-8-89176-256-5.
\newblock \doi{10.18653/v1/2025.findings-acl.583}.
\newblock URL \url{https://aclanthology.org/2025.findings-acl.583/}.

\bibitem[Zheng et~al.(2023)Zheng, Chiang, Sheng, Zhuang, Wu, Zhuang, Lin, Li,
  Li, Xing, Zhang, Gonzalez, and Stoica]{zheng2023judging}
Lianmin Zheng, Wei-Lin Chiang, Ying Sheng, Siyuan Zhuang, Zhanghao Wu, Yonghao
  Zhuang, Zi~Lin, Zhuohan Li, Dacheng Li, Eric~P. Xing, Hao Zhang, Joseph~E.
  Gonzalez, and Ion Stoica.
\newblock Judging llm-as-a-judge with mt-bench and chatbot arena.
\newblock In \emph{Proceedings of the 37th International Conference on Neural
  Information Processing Systems}, NIPS '23, Red Hook, NY, USA, 2023. Curran
  Associates Inc.
\newblock URL \url{https://dl.acm.org/doi/10.5555/3666122.3668142}.

\end{thebibliography}

\clearpage
\appendix

\section{Human validation of atomic \texttt{cite\_verify}}
\label{app:cite-verify-validation}

Strict accuracy is a deterministic gate on citation resolution and format, not on support. The component that verifies support is the atomic \texttt{cite\_verify} audit, itself an LLM output. We validate it against two law-trained annotators.

\paragraph{Protocol.} Two annotators independently labeled a stratified sample of $148$ atomic assertions drawn from the seed-$0$ audit traces, blind to the tool's verdict, assigning one of \texttt{SUPPORTED}/\texttt{PARTIAL}/\texttt{UNSUPPORTED}/\texttt{MISCITED} per assertion. These are the same two law-trained annotators as the judge calibration: research collaborators who consented to the use of their ratings and labeled through a blinded interface with the tool's verdict withheld. The sample carries an equal quota per tool-assigned class ($37$ each) so that minority-class recall is estimable. The annotators are not told this. The annotators agree with each other at $\kappa_{\mathrm{HH}}{=}0.846$ (unweighted four-class $\kappa$ between the two humans, exact agreement $88.5\,\%$), and we take their $131$-item agreed subset as gold.

\paragraph{Result.} As a \emph{binary} detector of claims needing attention (\texttt{PARTIAL}/\texttt{UNSUPPORTED}/\texttt{MISCITED} as the positive class) the tool reaches precision $0.802$, recall $0.875$, F1 $0.837$: it surfaces most claims a human would question, and most of what it flags is worth questioning. Its \emph{four-way} labels are weakly calibrated ($\kappa_{\mathrm{TH}}{=}0.275$ between tool and humans, exact agreement $45.8\,\%$): tool and annotators frequently agree that a claim is under-supported but disagree on which verdict applies, with \texttt{MISCITED} the least reliable class (recovering the gold label on $15$ of the $35$ \texttt{MISCITED}-assigned items that remain in the $131$-item agreed subset, from the $37$ originally sampled). The audit trace should therefore be read as a reliable flag carrying an advisory label, and the per-class verdict mix in \S\ref{sec:results} as indicative rather than precise. A verdict taxonomy the tool can apply reproducibly is the clearest next step. The correctness judge, by contrast, is calibrated ($r{=}0.85$, $\kappa_q{=}0.85$). Negative or absence claims, where no positive citation may exist, are a known edge case that future verdict definitions should handle explicitly. Three limits bound these numbers. Because annotators labeled the tool's own claim decomposition, the check cannot detect claims the splitter dropped or over-split, so the reported recall is an upper bound with respect to decomposition error. Validation also covers the verdict labels only, not the free-text reason each carries, so a label-correct trace with a flawed rationale is outside what these numbers certify. Finally, the two annotators are project collaborators rather than independent third parties, and the same pair supplies the judge calibration, so their labels may carry correlated expectation bias. An independent-annotator subset is left to future work.

\section{Why only one ablation is reported}
\label{app:ablation-identifiability}

\gandr's orchestrator returns the Round-1 draft whenever \texttt{protocol\_check} passes, and that return path does not read the Critic's verdict. Disabling the Critic therefore leaves the committed answer unchanged on every item that commits at Round~1, which is $98.0\,\%$ of runs for that variant and $98.6\,\%$ for full \gandr. The same holds for the holistic \texttt{cite\_verify} variant, which commits at Round~1 on $97.4\,\%$ of runs. The \mbox{$-$\,Critic} variant disables only the Critic's synthesis call, the step that emits the PASS/REWRITE verdict and any rewrite directives. The \texttt{protocol\_check}, atomic \texttt{cite\_verify}, and \texttt{sef\_rubric\_score} tools all still run, so the audit trace is still produced. Any accuracy difference they show is therefore a difference between sampling draws at temperature $0.7$, not an effect of the ablated component, and we do not report it as one. Removing the Round-1 anchor is different in kind: it commits at Round~1 on only $3.6\,\%$ of runs, sends most items through the full rewrite budget, and returns \texttt{passed=False} on $92.7\,\%$ of them, so its effect is attributable. Isolating the Critic and the decomposition mode would require a design in which they can change the committed text, which we leave to future work.

\section{Loop-depth distribution}
\label{app:loop-depth}

Across the $925$ \gandr runs ($k{=}5$, $T{=}3$, $185$ items), $912$ ($98.6\,\%$) commit on Round~1 via the protocol anchor, $0$ terminate at $t{=}2$, and $13$ ($1.4\,\%$) exhaust the rewrite budget, each on a single seed of a distinct item. No run reaches a passing answer through rewriting, so the loop's contribution is entirely the fail-closed exit: on exhaustion the orchestrator returns \texttt{passed=False} with the full audit trace, converting a silent commit into an auditable failure. It follows that a rewrite-disabled ($T{=}0$) variant would return the identical committed answer on all but these $13$ runs, which it would instead commit silently, so the loop's marginal effect on accuracy is zero by construction and its only effect is to convert those silent commits into auditable failures.

\section{\gandr prompts}
\label{app:gandr-prompts}

\gandr uses two prompts, one for the Drafter agent and one for the Critic agent. Each is rendered in a fresh LLM context per round, so the information barrier between Drafter and Critic is preserved.

\paragraph{Drafter system prompt.} Pins the CREAC five-block schema and the inline-citation contract. The illustrative mini-example shipped with the prompt is omitted here for space.

\begin{quote}
\small\ttfamily\raggedright
You are a legal analyst producing structured CREAC memos.\\[2pt]
You MUST emit your answer using exactly FIVE labeled blocks, in this order:\\[2pt]
\ [CONCLUSION]\ \ \ (one sentence, BLUF, no hedging)\\
\ [RULE]\ \ \ \ \ \ \ \ \ (applicable legal rule(s), with \{cite: <rule\_id>\} citations)\\
\ [EXPLANATION]\ \ (interpretation; every claim ends with \{cite: <rule\_id>\})\\
\ [APPLICATION]\ \ (apply rule to facts; every factual conclusion ends with \{cite: <rule\_id>\})\\
\ [CONCLUSION]\ \ \ (restate first conclusion, identical wording, one sentence)\\[2pt]
Five blocks. The closing [CONCLUSION] is mandatory. Without it the memo is rejected.\\[2pt]
Hard rules:\\
- Cite ONLY identifiers that appear verbatim as rule\_id headers in the RETRIEVED CONTEXT block. Use the exact rule\_id string shown in [<rule\_id>]; do NOT shorten, paraphrase, drop suffixes, or invent IDs.\\
- Every claim in [EXPLANATION] and [APPLICATION] must end with an in-line \{cite: <rule\_id>\}.\\
- No chain-of-thought, no scratch reasoning, no preamble. Begin immediately with [CONCLUSION].\\
- If the rule is ambiguous, surface the ambiguity inside [APPLICATION]; do NOT hedge in either [CONCLUSION] block.\\
- After the closing [CONCLUSION], stop. Do not add a postscript or summary.
\end{quote}

\paragraph{Drafter user template.} Renders the question, the retrieved passages (the only source of valid citations), and on rounds $t > 1$ the previous Critic's rewrite directives.

\begin{quote}
\small\ttfamily\raggedright
LEGAL QUESTION:\\
\{question\}\\[2pt]
RETRIEVED CONTEXT (rule\_id $\rightarrow$ text; this is the only source of valid citations):\\
\{context\}\\[2pt]
\{rewrite directives block, empty on Round 1\}\\[2pt]
Produce the CREAC memo now.
\end{quote}

\paragraph{Critic system prompt.} Runs in a separate LLM context with no access to the Drafter's reasoning. Emits a structured JSON verdict that lands in the audit trace.

\begin{quote}
\small\ttfamily\raggedright
You are a strict legal reviewer auditing a CREAC memo written by a junior associate.\\[2pt]
You will be given:\\
\ \ 1.\ The original legal question and the retrieved context.\\
\ \ 2.\ The draft CREAC memo to audit.\\
\ \ 3.\ Tool results (protocol\_check, cite\_verify, sef\_rubric\_score).\\[2pt]
You do NOT see the drafter's reasoning. You see only what a human reader would see.\\[2pt]
Your output MUST be valid JSON with this exact schema:\\
\ \ \{\\
\ \ \ \ "decision": "PASS" $|$ "REWRITE",\\
\ \ \ \ "rewrite\_directives": [ "concise, actionable instruction (one per line)" ],\\
\ \ \ \ "rationale": "one short paragraph explaining the decision"\\
\ \ \}\\[2pt]
Decision policy:\\
- PASS only if every tool report is green (protocol\_check passes, cite\_verify reports $\ge$ 95\% atomic assertions verified, sef\_rubric\_score mean $\ge$ 4.0 AND min $\ge$ 3.0).\\
- Otherwise REWRITE. Each directive must reference the specific failing assertion ID, CREAC block, or SEF dimension so the drafter can act on it without guessing.\\[2pt]
Examples of good directives:\\
- "Assertion 3 cited rule \S A but the cited span actually discusses \S B; either rewrite the claim to match \S B or remove it."\\
- "[APPLICATION] paragraph 2 contains an unsupported factual conclusion; add a \{cite: ...\} or remove the conclusion."\\
- "SEF dim AFL is 2/5: the opening [CONCLUSION] is hedged. Restate as a direct answer."\\[2pt]
Do not output anything except the JSON object.
\end{quote}

\paragraph{Critic user template.} Renders the question, the retrieved context, the draft under review, and the three tool reports.

\begin{quote}
\small\ttfamily\raggedright
LEGAL QUESTION:\\
\{question\}\\[2pt]
RETRIEVED CONTEXT (only valid source of citations):\\
\{context\}\\[2pt]
DRAFT CREAC MEMO TO AUDIT:\\
-{}-{}-\\
\{draft\}\\
-{}-{}-\\[2pt]
TOOL REPORTS:\\
- protocol\_check: \{protocol\_report\}\\
- cite\_verify (atomic-claim mode): \{cite\_verify\_report\}\\
- sef\_rubric\_score: \{sef\_report\}\\[2pt]
Produce your JSON decision now.
\end{quote}

\begin{sloppypar}\emergencystretch=3em
Note the deliberate gap between the Critic's stated decision policy and the orchestrator's behavior. The Critic is instructed to PASS only when every tool report is green, including a $95\,\%$ atomic-verification bar. The orchestrator's commit rule reads \texttt{protocol\_\allowbreak check} directly (the Round-1 anchor, \S\ref{sec:method}) rather than the Critic's PASS/REWRITE decision, so a Critic REWRITE verdict does not by itself trigger a rewrite. The Critic's decision and its directives are logged to the audit trace on every round regardless.
\end{sloppypar}

\section{Baseline prompt templates}
\label{app:baseline-prompts}

Every system in the cohort (B1--B5 and \gandr) carries the shared \textsc{grounding requirement} block in its primary system prompt. B2 inherits it from the \gandr Drafter prompt, and B4 and B5 carry it on their writing nodes (Writer for B4, Draft and Rewrite for B5). This parity is what makes strict scoring a content check rather than a compliance check.

\paragraph{Shared \textsc{grounding requirement} block (verbatim).}
\begin{quote}
\small\ttfamily\raggedright
GROUNDING REQUIREMENT (mandatory):\\
- Cite ONLY identifiers that appear verbatim as rule\_id headers in the RETRIEVED CONTEXT block (shown as [<rule\_id>] before each passage). Use the exact rule\_id string shown in [<rule\_id>]; do NOT shorten, paraphrase, drop suffixes, or invent IDs.\\
- Every factual or normative claim must be followed by an in-line \{cite: <rule\_id>\} marker referring to a rule\_id from the retrieved context.
\end{quote}

\paragraph{B1 zero-shot.}
\begin{quote}
\small\ttfamily\raggedright
You are a legal analyst. Answer the user's question concisely, drawing primarily from the retrieved context.\\[2pt]
\{shared GROUNDING REQUIREMENT block appended verbatim\}
\end{quote}

\paragraph{B2 SEF self-check, appended to the \gandr Drafter system prompt.}
\begin{quote}
\small\ttfamily\raggedright
SELF-CHECK BEFORE EMITTING (single shot, no second pass):\\
\ AFL\ \ state the answer in the FIRST [CONCLUSION] block AND restate it at the end.\\
\ AC\ \ \ the answer must be unambiguous (no ``it depends'' without a sub-conclusion).\\
\ CI\ \ \ the conclusion must be in its own block, separated from reasoning.\\
\ DTC\ \ use precise legal terms consistently (e.g.\ ``negligence'' not ``carelessness'').\\
\ CEA\ \ every claim in [APPLICATION] must end with a \{cite: rule\_id\}.\\
\ FS\ \ \ cite specific facts from the question, not vague generalities.
\end{quote}

\paragraph{B3 advanced RAG system prompt.}
\begin{quote}
\small\ttfamily\raggedright
You are a legal analyst. Use ONLY the retrieved context to answer.\\[2pt]
\{shared GROUNDING REQUIREMENT block appended verbatim\}
\end{quote}

\paragraph{B4 Researcher and Writer agents.} The two agents are configured verbatim as follows. The \textbf{Researcher} has role ``Legal Researcher,'' goal ``Identify the controlling legal rules that apply to the question,'' and backstory ``A meticulous law-firm research associate who never misattributes a citation.'' Its task description is ``Identify the controlling legal rules from the retrieved context.'' followed by the question and the retrieved context. The expected output is ``A list of relevant rule\_ids with one-line summaries.'' The \textbf{Writer} has role ``Legal Writer,'' goal ``Produce a clear, well-cited answer to the question,'' and backstory ``A senior associate who writes structured client memos,'' \emph{with the shared \textsc{grounding requirement} block appended to the backstory verbatim}. Its task description is ``Using the researcher's findings, write the final answer to the question.'' \emph{with the same \textsc{grounding requirement} block appended to the task description}, and its expected output is ``A complete legal answer with inline citations.'' The Researcher's output is passed as task context to the Writer through the CrewAI sequential pipeline.

\paragraph{B5 four-node prompts.} Each node sends a short instruction as the system message and a templated user message that includes the running graph state. The two writing-side nodes (Draft and Rewrite) carry the shared \textsc{grounding requirement} block.
\begin{quote}
\small\ttfamily\raggedright
Plan: ``You plan a legal answer. List 3-5 sub-questions to address.''\\[2pt]
Draft: ``You write the legal answer. Follow the plan.'' + \{GROUNDING REQUIREMENT\}\\[2pt]
Review: ``You critique your own draft. List concrete issues to fix, including any missing or malformed \{cite: <rule\_id>\} markers.''\\[2pt]
Rewrite: ``You rewrite the draft addressing the review.'' + \{GROUNDING REQUIREMENT\}
\end{quote}

\section{SEF rubric (Critic draft-quality scorer)}
\label{app:sef}

The \gandr Critic invokes a draft-quality scorer inherited from SEF~\citep{qian2026thinkingjustifyingaligninghighstakes}. The scorer rates the Drafter's CREAC memo on six $1$--$5$ dimensions and is called against the local generator at $T{=}0$. Its output appears in the audit trace as one tool report among three. It is \emph{not} consumed by any headline metric: lenient accuracy, strict accuracy, and Stable~\&~Correct~@~$5$ read no SEF dimension. The Critic uses the rubric only to phrase actionable rewrite directives when the protocol gate fails.

\begin{itemize}[itemsep=1pt]
\item \textbf{AFL (Answer First/Last).} The operative holding is stated in the opening \texttt{[CONCLUSION]} block \emph{and} restated identically in the closing one. A reader scanning only the first and last block should land on the same answer.
\item \textbf{AC (Answer Clarity).} The answer is unambiguous and direct. Hedges (``it depends'') without a sub-conclusion are penalized, because a verifier needs a holding to verify against.
\item \textbf{CI (Conclusion Isolation).} The conclusion is structurally separated from the reasoning, not embedded inside \texttt{[APPLICATION]}.
\item \textbf{DTC (Domain Terminology Consistency).} Legal terms are used precisely and consistently. Mixing ``negligence'' with ``carelessness'' as synonyms is penalized.
\item \textbf{CEA (Conclusion--Evidence Alignment).} Every claim in \texttt{[APPLICATION]} ends with an inline \texttt{\{cite: rule\_id\}} and the cited authority actually supports the claim.
\item \textbf{FS (Fact Specificity).} The answer cites specific facts from the question (party names, clauses, dates, amounts) rather than vague generalities.
\end{itemize}

\paragraph{SEF rubric prompt (verbatim).}
\begin{quote}
\small\ttfamily\raggedright
You are scoring an answer against the SEF (Structured Explanation Framework) rubric.\\[2pt]
Score each of the 6 dimensions on 1--5:\\
\ \ AFL (Answer First/Last):\ \ the answer is stated at the beginning AND end.\\
\ \ AC\ \ (Answer Clarity):\ \ \ \ the answer is unambiguous and direct.\\
\ \ CI\ \ (Conclusion Isolation):\ \ conclusion is structurally separated from reasoning.\\
\ \ DTC (Domain Terminology Consistency):\ \ legal terms are used precisely and consistently.\\
\ \ CEA (Conclusion-Evidence Alignment):\ \ the conclusion is grounded in cited evidence.\\
\ \ FS\ \ (Fact Specificity):\ \ the answer cites specific facts, not vague generalities.\\[2pt]
ANSWER TO SCORE:\\
\{candidate\}\\[2pt]
Output JSON only:\\
\{``AFL'': <1-5>, ``AC'': <1-5>, ``CI'': <1-5>, ``DTC'': <1-5>, ``CEA'': <1-5>, ``FS'': <1-5>\}
\end{quote}

\section{LLM-as-judge rubrics}
\label{app:judge-rubrics}

The judge is Claude Opus 4.7 (\texttt{claude-opus-4-7})~\citep{anthropic2026opus47}, invoked as a sub-agent. Both rubrics are reproduced exactly as rendered for every item.

\paragraph{Correctness rubric (1--5, headline).}
The judge sees the question, the ground-truth reference, and the candidate answer, and writes a single JSON verdict.

\begin{quote}
\small\ttfamily\raggedright
You are evaluating the CORRECTNESS of a legal answer.\\[2pt]
LEGAL QUESTION:\\
\{question\}\\[2pt]
GROUND TRUTH ANSWER (or reference solution):\\
\{ground\_truth\}\\[2pt]
CANDIDATE ANSWER:\\
\{candidate\}\\[2pt]
Score the candidate on a 1--5 scale where:\\
\ \ 5 = Reaches the same conclusion as ground truth, with sound reasoning and correct citations.\\
\ \ 4 = Reaches the same conclusion; minor reasoning gaps or one weak citation.\\
\ \ 3 = Reaches the same conclusion but with notable reasoning errors or 2+ weak citations.\\
\ \ 2 = Wrong conclusion but partially correct reasoning.\\
\ \ 1 = Wrong conclusion AND wrong reasoning, or refuses to answer.\\[2pt]
Special case --- binary tasks. The 1--5 rubric does not apply gradationally when the GROUND TRUTH ANSWER is one of:\\
\ \ (a) a yes/no/true/false token (case-insensitive, possibly with trailing punctuation: ``Yes'', ``No.'', ``True'', ``False''),\\
\ \ (b) a single-rule citation in the form ``\S\ N'', ``Section N'', ``Rule N'', or a known statute reference like ``28 U.S.C.\ \S\ 1332'', ``Rule 23''.\\
For (a) and (b), score 5 if the candidate's operative holding matches the ground truth (regardless of phrasing or intermediate reasoning), else score 1. Do not use intermediate scores 2/3/4 for these items.\\[2pt]
Numerical-answer tasks (e.g., a dollar amount like ``\$187552'' or ``\$3,390'') are NOT binary --- close-enough matters. Apply the 1--5 rubric: score 5 if the candidate produces the exact figure, 4 if within $\pm 1\,\%$ (a rounding artifact), 3 if within $\pm 10\,\%$, 2 if the right order of magnitude but materially off, 1 if fundamentally wrong or refused.\\[2pt]
Open-ended tasks where the ground truth is a long evidence span (e.g., LegalBench-RAG contract-NLI items: question ``Does the document permit X?'', gold = a quoted clause) --- apply the 1--5 rubric to whether the candidate's operative \emph{yes/no holding} aligns with what the gold clause supports.\\[2pt]
Output JSON only:\\
\{``score'': <1-5>, ``rationale'': ``<one sentence>''\}
\end{quote}

\paragraph{Conclusion-equivalence rubric (S\&C@$k$).}
Used for the $C(5,2){=}10$ pairwise judgments per item that feed Stable~\&~Correct~@~$5$. The judge sees the question and two candidate answers in random order, system identity stripped.

\begin{quote}
\small\ttfamily\raggedright
You are evaluating whether two CREAC-style legal answers reach the SAME LEGAL CONCLUSION.\\[2pt]
Two answers are EQUIVALENT if a legal practitioner would treat them as making the same operative holding, even if they use different phrasing, cite different authorities, or differ in detail. They are NOT equivalent if they (a) reach a different legal disposition, (b) apply a different legal test, or (c) disagree on the controlling rule.\\[2pt]
Examples (illustrative only --- apply judgment to the actual answers):\\
\ \ EQUIVALENT:\ \ ``the test is the nerve center test'' $\leftrightarrow$ ``principal place of business is the headquarters where high-level officers direct the company''\\
\ \ NOT EQUIV.:\ \ ``diversity jurisdiction is codified in 28 U.S.C.\ \S\ 1332'' $\leftrightarrow$ ``diversity jurisdiction is codified in \S\S\ 1331 and 1332'' (the second adds a wrong statute)\\[2pt]
LEGAL QUESTION:\\
\{question\}\\[2pt]
ANSWER A:\\
-{}-{}-\\
\{answer\_a\}\\
-{}-{}-\\[2pt]
ANSWER B:\\
-{}-{}-\\
\{answer\_b\}\\
-{}-{}-\\[2pt]
Output JSON only:\\
\{``equivalent'': true $|$ false, ``rationale'': ``<one sentence>''\}
\end{quote}

\paragraph{Sub-agent execution rules.}
The sub-agent receives the two rubrics above inside a longer instruction template that pins behavioral rules: (i) read the entire request batch before writing any verdicts, (ii) apply the rubric literally without introducing additional criteria, (iii) penalize hallucinated citations even when the prose is fluent, (iv) preserve each \texttt{item\_id} verbatim, (v) emit one valid-JSON verdict per line with no surrounding markdown, and (vi) write the verdict file in a single write call rather than incrementally. Two further rubrics live in the same template as offline development probes: citation faithfulness (per-citation grounding against the retrieved context) and a blinded pairwise A/B comparing two outputs for verification speed. Neither is consumed by any headline metric.

\section{System pipelines}
\label{app:pipelines}

Every system consumes the same benchmark item and the same BM25 top-$8$ retrieval context. The systems differ only in what they do with that input.

\paragraph{\gandr (B6, ours).}
\begin{sloppypar}\emergencystretch=3em
Two-agent loop. Each round repeats four steps. (1)~The Drafter is called with the system prompt and user template of Appendix~\ref{app:gandr-prompts}, producing a CREAC-structured draft. (2)~Three tools run on the draft: the regex \texttt{protocol\_\allowbreak check}, the atomic-claim \texttt{cite\_\allowbreak verify} at $T{=}0$, and the six-dimension \texttt{sef\_\allowbreak rubric\_\allowbreak score} at $T{=}0$. (3)~The Critic is called in a separate LLM context with the question, the retrieved context, the draft, and the three tool reports, emitting a JSON PASS/REWRITE verdict that is appended to the audit trace. (4)~The orchestrator's commit rule is anchored to \texttt{protocol\_\allowbreak check}. If the structural check passes, the orchestrator commits Round~1 and the Critic acts as auditor. Otherwise the Critic's rewrite directives are prepended to the Drafter's next user prompt and the loop iterates up to $T{=}3$ rounds. On loop exhaustion the system returns the last draft with \texttt{passed=\allowbreak False} rather than falling through to a best-effort PASS.
\end{sloppypar}

\paragraph{B1 zero-shot.} Single LLM call. The system prompt is the role assignment plus the shared \textsc{grounding requirement} block. The user prompt concatenates the question and the top-$8$ retrieved passages. No CREAC schema, no critique, no agent loop.

\paragraph{B2 SEF-prompted single LLM.} Single LLM call. The system prompt is the \gandr Drafter system prompt with the six-line SEF self-check appended. User prompt and retrieval are identical to B1. No Critic and no rewrite loop. B2 is the strongest baseline and the head-to-head comparator.

\paragraph{B3 advanced RAG.} BM25 retrieves a top-$20$ candidate pool, a cross-encoder re-ranker (BAAI \texttt{bge-reranker-large}, with Cohere \texttt{rerank-3.5} as a configured fallback) re-orders the pool, and the top $5$ passages are passed to a single LLM call carrying the shared \textsc{grounding requirement} block. No CREAC schema. B3 varies retrieval rather than generation, which is what makes the B3-vs-\gandr contrast a test of whether the lead could be a retrieval-quality artifact.

\paragraph{B4 CrewAI (Researcher + Writer).} Two-agent sequential pipeline~\citep{crewai2024}. The Researcher returns a list of controlling \texttt{rule\_id} strings with one-line summaries. The Writer produces the final answer with inline \texttt{\{cite: <rule\_id>\}} markers using the Researcher's output as context. Both agents run against the shared backbone. No Critic, no atomic-claim verifier, no rewrite loop.

\paragraph{B5 LangGraph self-correction.} A four-node graph~\citep{langgraph2024} that runs in a \emph{single} LLM context: (1)~Plan ($3$--$5$ sub-questions), (2)~Draft (answer with inline cites), (3)~Review (critique the just-written draft), (4)~Rewrite (final answer addressing the review). Every node reads and writes the same graph state, so the review is computed in the same context that produced the draft. B5-vs-\gandr is the single-context versus separate-context contrast.

\section{Benchmark and retrieval construction}
\label{app:dataset}

\paragraph{LegalBench.} The benchmark draws $15$ LegalBench~\citep{guha2023legalbench} task tags: \texttt{hearsay}, \texttt{learned\_hands} sub-tasks, \texttt{nys\_judicial\_ethics}, \texttt{definition\_extraction}, \texttt{overruling}, \texttt{scalr}, \texttt{ssla}, \texttt{rule\_qa}, \texttt{citation\_prediction}, \texttt{consumer\_contracts\_qa}, \texttt{corporate\_lobbying}, \texttt{personal\_jurisdiction}, and \texttt{maud}.

\paragraph{LegalBench-RAG.} The contract-NLI and CUAD subsets~\citep{pipitone2024legalbenchrag} populate the \texttt{cross\_domain} bucket and provide open-ended grounding-required questions against contract clauses.

\paragraph{Retrieval index.} A single BM25 index over $13{,}090$ passages. The corpus merges two sources. First, every inline context block from a LegalBench item is split into rule-keyed passages of the form (\texttt{rule\_id}, \texttt{text}). Second, the LegalBench-RAG retrieval mirror is concatenated, preserving the original \texttt{rule\_id} strings. Tokenization is the standard word-level pattern, lower-cased. The implementation is \texttt{rank\_bm25.BM25Okapi} with default $k_1$ and $b$. The index is built once and reused across all systems, seeds, and $k$-th repeats, which is the invariant that makes Stable~\&~Correct~@~$k$ measure system behavior rather than retrieval randomness.

\paragraph{Per-item schema.} Each item is a JSON record with \texttt{item\_id} (the join key across run artifacts), \texttt{question}, \texttt{ground\_truth}, \texttt{domain} (one of eight buckets), \texttt{reasoning\_type} (\texttt{statutory}, \texttt{case-law}, or \texttt{mixed}), \texttt{source}, \texttt{is\_yesno}, and \texttt{extra}.

\paragraph{Licenses and intended use.} LegalBench is MIT-licensed and LegalBench-RAG is CC-BY-4.0. Both are distributed for legal-NLP research evaluation and permit research redistribution. \texttt{rank\_bm25} is Apache-2.0, the bge-reranker-large weights are MIT, and CrewAI and LangGraph are both MIT. The Nemotron-Cascade-2-30B-A3B FP8 quantization~\citep{nvidia2026nemotroncascade2} is distributed under the NVIDIA Open Model License, GLM-4.7-Flash~\citep{zai2026glm47flash} under its publisher's terms, and \texttt{gpt-5.4} and \texttt{gpt-5.4-mini}~\citep{openai2026gpt54} are accessed under the OpenAI API Terms of Service. Our use of every artifact is consistent with its license's research scope.

\section{Protocol check and strict correctness}
\label{app:strict}

\paragraph{CREAC \texttt{protocol\_check}.} A regex parser, with no LLM call, verifies that the five labeled blocks \texttt{[CONCLUSION]}, \texttt{[RULE]}, \texttt{[EXPLANATION]}, \texttt{[APPLICATION]}, \texttt{[CONCLUSION]} appear in order and that both conclusion blocks are non-empty.

\paragraph{Grounding check (cite-validity).} The answer text is scanned for inline markers of the form \texttt{\{cite:\,<rule\_id>\}}. An answer is \emph{grounded} iff (i)~at least one such marker is present, \emph{and} (ii)~every extracted marker resolves to a \texttt{rule\_id} appearing verbatim (exact substring match, with the \texttt{corpus:} prefix stripped on either side) in some passage of the item's retrieved top-$8$ context. A single fabricated or out-of-context marker fails the check, even when every other marker in the same answer resolves correctly.

\paragraph{Strict correctness.} An item is strict-correct iff the judge correctness score is $\ge 4$ \emph{and} the grounding check passes. Lenient correctness drops the grounding conjunct.

\paragraph{Atomic \texttt{cite\_verify}.} For \gandr the Critic additionally invokes an atomic-claim decomposition tool that splits the \texttt{[EXPLANATION]} and \texttt{[APPLICATION]} blocks into per-assertion tuples (\texttt{claim}, \texttt{rule\_id}, \texttt{supporting\_span}) and labels each \texttt{SUPPORTED}, \texttt{PARTIAL}, \texttt{UNSUPPORTED}, or \texttt{MISCITED} at $T{=}0$. These labels populate the audit trace on every round. The strict column does \emph{not} consume this tool; it consumes only the grounding check above. The human validation of these labels is reported in Appendix~\ref{app:cite-verify-validation}.

\section{Models and runtime settings}
\label{app:models}

\paragraph{Generator backbones.} The headline run is on NVIDIA Nemotron-Cascade-2-30B-A3B~\citep{nvidia2026nemotroncascade2}, a Mamba2-Transformer hybrid MoE with $30$B total and $3$B active parameters per token, served via vLLM in FP8 on a single H100 NVL. The pluggability check additionally exercises GLM-4.7-Flash~\citep{zai2026glm47flash} (a 30B/3B-active MoE from a different open-source family, same vLLM stack) and two commercial models reached via the OpenAI API, \texttt{gpt-5.4-mini} and \texttt{gpt-5.4}~\citep{openai2026gpt54}. A provider-abstraction client keeps the orchestrator code unchanged across backbones; the OpenAI branch uses \texttt{max\_completion\_tokens} in place of \texttt{max\_tokens} and skips the vLLM-specific thinking flag.

\paragraph{Generation hyperparameters.} All systems use generator temperature $0.7$ and \texttt{max\_tokens}$\,{=}\,2048$ on the Drafter or single-LLM call. The atomic \texttt{cite\_verify} and \texttt{sef\_rubric\_score} tools inside the \gandr Critic call the local generator at $T{=}0$. The Critic's own LLM call uses temperature $0.7$. For repeated runs ($k{=}5$), seeds $\{0,1,2,3,4\}$ are passed through to vLLM. The thinking flag is \texttt{false} on every system, so no model emits a reasoning preface.

\paragraph{Loop bounds.} $T{=}3$ maximum Critic rounds. The atomic-verification threshold inside the rewrite path is $0.50$, a corpus-calibrated value. The Round-1-anchor ablation reverts to the original $0.95$ gate and shows that the protocol-anchored commit, which does not depend on the threshold, is the load-bearing choice.

\section{Extended judge calibration}
\label{app:calibration}

\begin{figure}[!htb]
\centering
\includegraphics[width=0.85\linewidth]{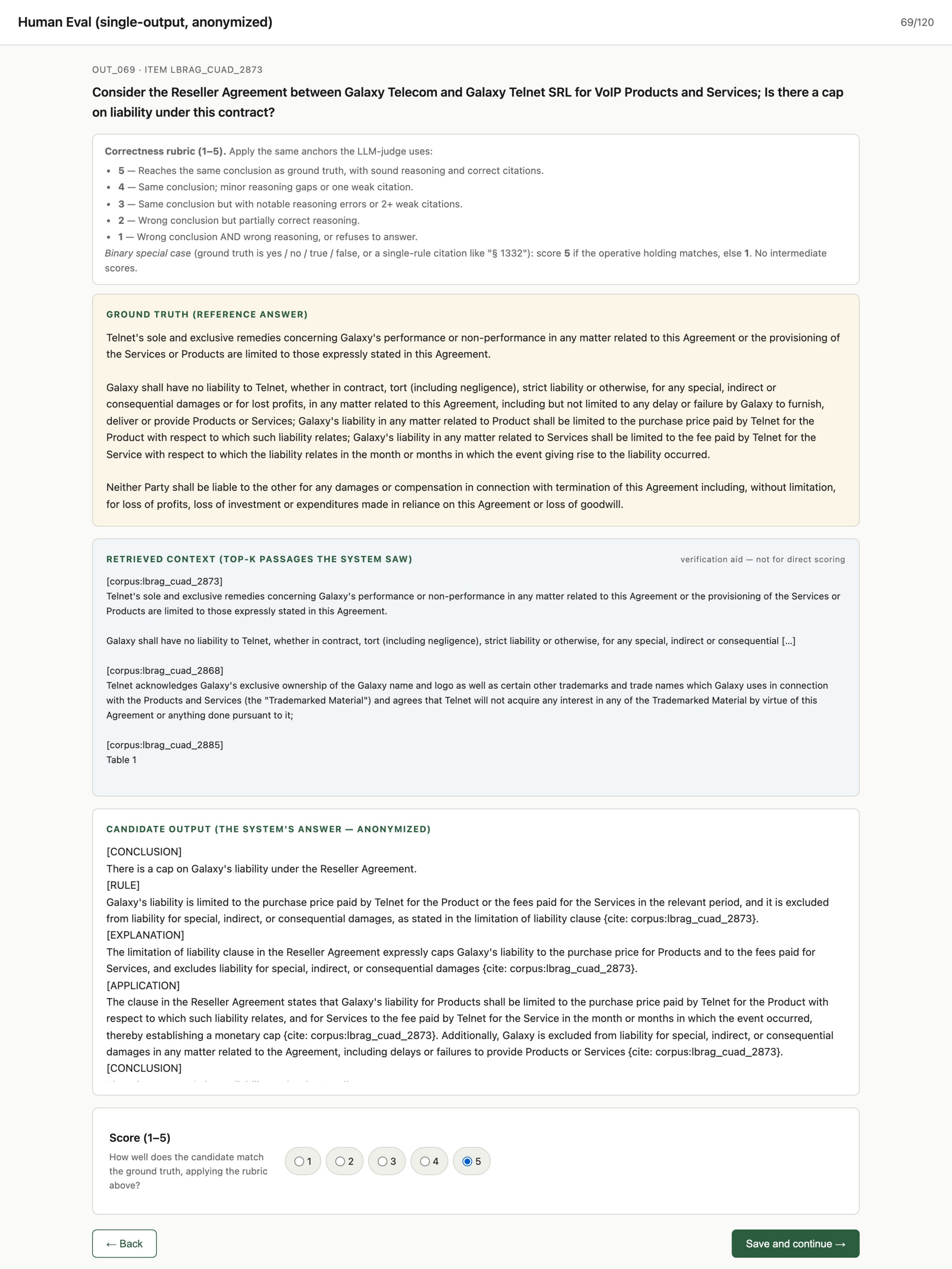}
\caption{Blinded annotation interface used for the $120$-item judge calibration. Each screen shows the question, the ground-truth reference answer, the top-$K$ retrieved context, and the anonymized candidate output. System identity and the judge's score are stripped, so an annotator cannot infer which system produced the answer or what the judge decided.}
\label{fig:annotation-interface}
\end{figure}

The $120$-item calibration set is $20$ items per system, stratified by judge correctness band. Both annotators clear the pre-registered thresholds of $r \ge 0.80$ and $\kappa_q \ge 0.60$ (pooled $r{=}0.846$, $\kappa_q{=}0.845$), and they agree with each other at $\kappa_q{=}0.979$. The annotators are research collaborators who consented to the use of their ratings and rated through a blinded interface with system identity and judge score stripped (Figure~\ref{fig:annotation-interface}).

\begin{table}[!htb]
\small
\centering
\setlength{\tabcolsep}{5pt}
\caption{Per-system calibration breakdown ($n{=}20$ per system). Mean columns report mean ratings by annotators A1, A2, and the judge. Both $r$ and $\kappa_q$ are reported per annotator against the judge. Every system clears the per-system $\kappa_q \ge 0.60$ threshold, so no single system drives the pooled agreement.}
\begin{tabular}{lccccccc}
\toprule
\textbf{System} & \textbf{A1} & \textbf{A2} & \textbf{Judge} & \textbf{$r$(A1)} & \textbf{$r$(A2)} & \textbf{$\kappa_q$(A1)} & \textbf{$\kappa_q$(A2)} \\
\midrule
B1 zero-shot     & 3.15 & 3.15 & 3.50 & 0.78 & 0.78 & 0.76 & 0.76 \\
B2 SEF prompt    & 3.15 & 3.15 & 2.90 & 0.89 & 0.89 & 0.88 & 0.88 \\
B3 advanced RAG  & 3.30 & 3.35 & 3.30 & 0.99 & 0.99 & 0.99 & 0.99 \\
B4 CrewAI        & 2.70 & 2.70 & 2.50 & 0.86 & 0.86 & 0.85 & 0.85 \\
B5 LangGraph     & 3.55 & 3.55 & 2.80 & 0.86 & 0.86 & 0.78 & 0.78 \\
B6 \gandr        & 2.85 & 2.70 & 3.05 & 0.86 & 0.75 & 0.85 & 0.74 \\
\bottomrule
\end{tabular}
\label{tab:per-system-calib}
\end{table}

\begin{table}[!htb]
\small
\centering
\setlength{\tabcolsep}{5pt}
\caption{Calibration by the judge's own correctness band ($n{=}120$). Exact and within-one agreement is high at both rubric ends ($1$ and $5$). Agreement collapses on the 4-band, where both annotators round up to $5$ on every item the judge rates $4$. The pattern is a rubric-anchor disagreement on the specific $4 \to 5$ transition rather than a systematic bias across the scale, and collapsing $\{4,5\}$ lifts exact agreement to $86\,\%$ without changing any pass/fail decision.}
\begin{tabular}{cccccccc}
\toprule
\textbf{Judge} & \textbf{$n$} & \textbf{A1} & \textbf{A2} & \textbf{A1$={}$J} & \textbf{A2$={}$J} & \textbf{$|\Delta_{A1}|{\le}1$} & \textbf{$|\Delta_{A2}|{\le}1$} \\
\midrule
1 & 50 & 1.28 & 1.30 & 86\,\% & 84\,\% & 92\,\%  & 92\,\%  \\
2 & 6  & 2.83 & 2.83 & 50\,\% & 50\,\% & 67\,\%  & 67\,\%  \\
3 & 6  & 2.67 & 2.67 & 50\,\% & 50\,\% & 67\,\%  & 67\,\%  \\
4 & 9  & 5.00 & 5.00 & 0\,\%  & 0\,\%  & 0\,\%   & 0\,\%   \\
5 & 49 & 4.73 & 4.67 & 92\,\% & 92\,\% & 100\,\% & 100\,\% \\
\bottomrule
\end{tabular}
\label{tab:per-judge-bin}
\end{table}

\section{Cost accounting}
\label{app:cost}

Each Critic round costs $4+N$ LLM calls, where $N$ is the atomic-assertion count: Drafter, \texttt{cite\_verify} decompose, $N$ per-claim verifies, \texttt{sef\_rubric\_score}, and Critic synthesis. Because the Round-1 anchor commits most items in a single round, the dollar factor over B2 is far smaller than the call-count factor.

\begin{table}[!htb]
\small
\centering
\setlength{\tabcolsep}{6pt}
\renewcommand{\arraystretch}{1.05}
\caption{Order-of-magnitude cost on the closed-source panes. The per-round breakdown enumerates every LLM call inside one Critic round, and the per-call mean for \texttt{cite\_verify} verify reflects the empirical atomic-assertion count on each backbone. Dollar figures use OpenAI list prices (2026-05-22) with per-call token usage from the $k{=}5$ artifacts. Self-hosted backbones incur no per-token charge.}
\begin{tabular}{@{}p{0.48\textwidth}cc@{}}
\toprule
\textbf{Component} & \textbf{\texttt{gpt-5.4-mini}} & \textbf{\texttt{gpt-5.4}} \\
\midrule
\multicolumn{3}{@{}l}{\textit{Per-round LLM-call breakdown (\gandr Critic)}} \\
\quad Drafter                              & 1             & 1              \\
\quad \texttt{cite\_verify} decompose      & 1             & 1              \\
\quad \texttt{cite\_verify} verify (mean $N$) & 5.4        & 8.6            \\
\quad \texttt{sef\_rubric\_score}          & 1             & 1              \\
\quad Critic synthesis                     & 1             & 1              \\
\quad \textbf{Total per round} ($4{+}N$)   & \textbf{9.4}  & \textbf{12.6}  \\
\midrule
\multicolumn{3}{@{}l}{\textit{Per-item LLM calls (averaged over $k{=}5$)}} \\
\quad B2 SEF                               & 1.00          & 1.00           \\
\quad \gandr                               & 9.78          & 13.03          \\
\quad Ratio \gandr / B2                    & 9.8$\times$   & 13.0$\times$   \\
\midrule
\multicolumn{3}{@{}l}{\textit{Estimated cost, $185$-item benchmark, single seed}} \\
\quad B2 SEF                               & $\sim$\$0.85  & $\sim$\$3.10   \\
\quad \gandr                               & $\sim$\$2.45  & $\sim$\$10.30  \\
\bottomrule
\end{tabular}
\label{tab:closed-source-cost}
\end{table}

\gandr on \texttt{gpt-5.4-mini} ($\sim$\$2.45) reaches $77.8\,\%$ strict accuracy, above B2 on \texttt{gpt-5.4} ($\sim$\$3.10, $76.8\,\%$) and within $2.2$\,pp of \gandr on the flagship at roughly a quarter of its cost.

\section{Baseline failure taxonomy}
\label{app:failure}

The four non-CREAC baselines fail cite-validity in qualitatively different ways under the same \textsc{grounding requirement} block, so the differences reflect what each system does \emph{with} the instruction rather than whether it received it.

\begin{table}[!htb]
\small
\centering
\setlength{\tabcolsep}{6pt}
\caption{Strict-lenient drops for the four non-CREAC, non-\gandr baselines, with the dominant failure mode for each. All four carry the same \textsc{grounding requirement} block.}
\begin{tabular}{@{}lc>{\raggedright\arraybackslash}p{7cm}@{}}
\toprule
\textbf{System} & \textbf{$\Delta$ strict} & \textbf{Dominant failure mode} \\
\midrule
B1 zero-shot     & $-18.9$\,pp & no canonical marker emitted \\
B3 advanced RAG  & $-17.3$\,pp & no marker plus fabricated-id residual \\
B4 CrewAI        & $-36.8$\,pp & fabricated \texttt{rule\_id} (invented label) \\
B5 LangGraph     & $-29.8$\,pp & \texttt{rule\_id} corruption (suffix, brackets) \\
\bottomrule
\end{tabular}
\label{tab:b1-cite-form}
\end{table}

\paragraph{B1 zero-shot ($-18.9$\,pp).} Emits no canonical marker on a sub-population of items, using non-ASCII brackets, square brackets, or braces without the \texttt{cite:} prefix. The disposition is often right while the citation form is unparseable.

\paragraph{B3 advanced RAG ($-17.3$\,pp).} Shows the same no-marker mode plus a residual of canonical markers pointing at a \texttt{rule\_id} outside the retrieved top-$8$. Re-ranking improves which passages arrive but does not change how the model cites them.

\paragraph{B4 CrewAI ($-36.8$\,pp).} The Researcher hands the Writer doctrine-label identifiers that never existed in the corpus, and the Writer cites them through a well-formed marker. The fabrication is introduced by the extra agent hop, which is why more agent calls without an audit amplify rather than reduce the failure.

\paragraph{B5 LangGraph ($-29.8$\,pp).} Corrupts retrieved identifiers with appended subsection suffixes and nested brackets, and the single-context Review node ratifies the corruption it just produced. This is the concrete mechanism behind the separate-context argument in \S\ref{sec:method}.

\section{Per-domain significance and marker normalization}
\label{app:per-domain}

\paragraph{Per-domain strict accuracy with CIs and McNemar.}
Table~\ref{tab:per-domain} reports per-domain strict accuracy for \gandr and B2 with $95\,\%$ Wilson intervals and paired McNemar $(b,c,p)$, computed under the strict criterion used throughout the paper. The per-domain rows sum to the aggregate ($b{=}38$, $c{=}17$, $p{=}0.006$) reported in \S\ref{sec:results} and in Table~\ref{tab:mcnemar}. At $n\approx 20$--$30$ the CIs are wide and overlap almost everywhere, and only employment separates the systems on its own, so the aggregate carries the claim. B2 is the pre-specified primary comparison, so its $p{=}0.006$ is the head-to-head test; it holds at $0.03$ under a conservative Bonferroni correction across all five baselines, and the four non-B2 comparisons clear any such correction at $p<10^{-8}$.

\begin{table}[htb]
\small
\centering
\setlength{\tabcolsep}{5pt}
\caption{Per-domain strict accuracy, \gandr (B6) vs.\ B2, with $95\,\%$ Wilson CIs and paired McNemar $(b,c,p)$, computed under the single strict criterion used throughout the paper. The ALL row reproduces Table~\ref{tab:headline} and Table~\ref{tab:mcnemar} exactly.}
\begin{tabular}{lccccc}
\toprule
\textbf{domain} & \textbf{$n$} & \textbf{B6 [CI]} & \textbf{B2 [CI]} & \textbf{$(b,c)$} & \textbf{$p$} \\
\midrule
constitutional     & 20 & 55.0 [34,74] & 80.0 [58,92] & (0,5) & 0.062 \\
contract           & 25 & 92.0 [75,98] & 76.0 [57,89] & (4,0) & 0.125 \\
criminal proc.     & 20 & 75.0 [53,89] & 50.0 [30,70] & (7,2) & 0.180 \\
cross\_domain      & 30 & 46.7 [30,64] & 43.3 [27,61] & (4,3) & 1.000 \\
employment         & 20 & 100 [84,100] & 45.0 [26,66] & (11,0) & \textbf{0.001} \\
federal proc.      & 25 & 48.0 [30,67] & 36.0 [20,55] & (6,3) & 0.508 \\
securities         & 20 & 85.0 [64,95] & 90.0 [70,97] & (1,2) & 1.000 \\
tort               & 25 & 76.0 [57,89] & 64.0 [45,80] & (5,2) & 0.453 \\
\midrule
\textbf{ALL}       & 185 & \textbf{70.8 [64,77]} & \textbf{59.5 [52,66]} & \textbf{(38,17)} & \textbf{0.006} \\
\bottomrule
\end{tabular}
\label{tab:per-domain}
\end{table}

\begin{table}[htb]
\small
\centering
\setlength{\tabcolsep}{8pt}
\caption{McNemar exact-test triples for \gandr versus each baseline. $b$: \gandr correct, baseline wrong; $c$: baseline correct, \gandr wrong.}
\begin{tabular}{lcc}
\toprule
\textbf{vs.\ baseline} & \textbf{Strict $(b,c,p)$} & \textbf{S\&C $(b,c,p)$} \\
\midrule
B1 zero-shot   & $(67,7, 2.1{\times}10^{-13})$ & $(78,2, 5.4{\times}10^{-21})$ \\
B2 SEF prompt  & $(38,17, 0.006)$              & $(39,9, 1.5{\times}10^{-5})$  \\
B3 adv.\ RAG   & $(66,15, 8.6{\times}10^{-9})$ & $(77,1, 5.2{\times}10^{-22})$ \\
B4 CrewAI      & $(100,9, 1.4{\times}10^{-20})$ & $(85,0, 5.2{\times}10^{-26})$ \\
B5 LangGraph   & $(98,7, 1.2{\times}10^{-21})$ & $(85,0, 5.2{\times}10^{-26})$ \\
\bottomrule
\end{tabular}
\label{tab:mcnemar}
\end{table}

\paragraph{Marker-normalization control.}
Because several of the baselines fail cite-validity through marker \emph{form} (Appendix~\ref{app:failure}), we ask what remains when form is forgiven. We re-score the stored outputs of all six systems, changing nothing about generation, under a normalization that folds non-ASCII bracket families to ASCII, treats the \texttt{cite:} prefix as optional whenever the bracketed content resolves, and matches corrupted identifiers by stem so that appended subsection suffixes still resolve. An explicit citation attempt that resolves to nothing is still counted as a failure, since normalizing form cannot repair a fabricated source. Strict accuracy then rises to $56.2\,\%$ for B1, $60.0\,\%$ for B2, $60.0\,\%$ for B3, $34.6\,\%$ for B4, and $38.4\,\%$ for B5, against $71.9\,\%$ for \gandr. Measured against each system's lenient score, the residual separates the cohort into two regimes. For B1 and B3 it closes to $1.1$ and $0.5$\,pp, so their strict collapse was almost entirely formatting, and for those two systems the strict criterion was largely measuring marker form. For B4 and B5 large residuals survive, $23.8$ and $13.0$\,pp, and no normalization can remove them: these are content-level failures, the fabricated doctrine labels and corrupted identifiers catalogued in Appendix~\ref{app:failure}, where the citation points at no retrievable source rather than at the right source in the wrong notation. That residual is the criterion behaving as intended, since it shows strict accuracy penalising invented provenance and not merely unconventional punctuation. \gandr's lead is likewise not an artifact of form: it survives normalization at $+11.9$\,pp over the strongest baseline, marginally wider than the $+11.3$\,pp headline gap, because what remains once form is forgiven is a difference in how often the answer is right.

\end{document}